\documentclass[10pt,twocolumn,letterpaper]{article}

\usepackage{cvpr}              % To produce the CAMERA-READY version
\usepackage[dvipsnames]{xcolor}

\usepackage[pagebackref,breaklinks,colorlinks,citecolor=cyan]{hyperref}

\usepackage{multirow}
\usepackage{tabularx}
\usepackage{booktabs}
\usepackage{bbding}
\usepackage{amssymb}% http://ctan.org/pkg/amssymb
\usepackage[accsupp]{axessibility} % Improves PDF readability for those with visual impairments.

\usepackage{pifont}% http://ctan.org/pkg/pifont
\usepackage{algorithm}
\usepackage[noend]{algpseudocode}

\usepackage{bm}

\usepackage{color, colortbl}

\usepackage[dvipsnames]{xcolor}

\definecolor{increase}{HTML}{FCE4D6}
\definecolor{decrease}{HTML}{FCE4D6}

\definecolor{FFABA8}{HTML}{e3ebfc}
\colorlet{Light}{FFABA8}
\newcommand{\CC}[1]{\cellcolor{Light}}

\definecolor{f5f8ff}{HTML}{f5f8ff}
\colorlet{Light2}{f5f8ff}
\newcommand{\CCC}[1]{\cellcolor{Light2}}

\definecolor{b4b4fa}{HTML}{d7d7fa}
\colorlet{Dark}{b4b4fa}

\def\paperID{7547} % *** Enter the Paper ID here
\def\confName{CVPR}
\def\confYear{2024}

\def\modelname{\textsc{MeLFusion}\xspace}
\def\ourdataset{MeLBench\xspace}  %MusicScribe
\def\ourdatasetsize{11,250\xspace}

\def\imagemusicmetric{IMSM\xspace}
\def\musiccapssamples{7,684\xspace}

\newcommand{\customsubsection}[1]{%
  \par
  \pagebreak[2]%
  \refstepcounter{subsection}%
    \everypar={%
      {\setbox0=\lastbox}% Remove the indentation
      \addcontentsline{toc}{subsection}{%
        {\protect\makebox[0.3in][r]{\thesubsubsection.} \hspace*{3pt}#1}}%
      \textbf{\thesubsection\space\space{#1}\space}%
      \everypar={}%
    }%
  \ignorespaces
}

\newcommand{\customsubsubsection}[1]{%
  \par
  \pagebreak[2]%
  \refstepcounter{subsubsection}%
    \everypar={%
      {\setbox0=\lastbox}% Remove the indentation
      \addcontentsline{toc}{subsubsection}{%
        {\protect\makebox[0.3in][r]{\thesubsubsection.} \hspace*{3pt}#1}}%
      \textbf{\thesubsubsection\space\space{#1}\space}%
      \everypar={}%
    }%
  \ignorespaces
}

\newcommand\blfootnote[1]{%
  \begingroup
  \renewcommand\thefootnote{}\footnote{#1}%
  \addtocounter{footnote}{-1}%
  \endgroup
}

\title{\modelname: Synthesizing \underline{M}usic from Imag\underline{e} and \underline{L}anguage Cues using \\ Dif\underline{fusion} Models}
\author{Sanjoy Chowdhury$^{*1,3 \dagger}$ $\quad$ Sayan Nag$^{*2,3 \dagger}$ $\quad$ K J Joseph$^3$ \\ Balaji Vasan Srinivasan$^3$ $\quad$ Dinesh Manocha$^1$ \vspace{2mm} \\ 
$^{1}$University of Maryland, College Park $\quad$ $^{2}$University of Toronto  $\quad$ $^{3}$Adobe Research\vspace{2mm}\\ 
\tt\small \{sanjoyc,dmanocha\}@umd.edu $\quad$ sayan.nag@mail.utoronto.ca $\quad$ \tt\small \{josephkj,balsrini\}@adobe.com $\quad$ \\
\tt\small Project page -\ \url{https://schowdhury671.github.io/melfusion_cvpr2024/}}

\begin{document}
\maketitle
\begin{abstract}
\label{abstract}

Music is a universal language that can communicate emotions and feelings. It forms an essential part of the whole spectrum of creative media, ranging from movies to social media posts. Machine learning models that can synthesize music are predominantly conditioned on textual descriptions of it. Inspired by how musicians compose music not just from a movie script, but also through visualizations, we propose \modelname, a model that can effectively use cues from a textual description and the corresponding image to synthesize music. \modelname is a text-to-music diffusion model with a novel ``visual synapse", which effectively infuses the semantics from the visual modality into the generated music. To facilitate research in this area, we introduce a new dataset \ourdataset, and propose a new evaluation metric \imagemusicmetric. Our exhaustive experimental evaluation suggests that adding visual information to the music synthesis pipeline significantly improves the quality of generated music, measured both objectively and subjectively, with a relative gain of up to \textbf{67.98\%} on the FAD score. We hope that our work will gather attention to this pragmatic, yet relatively under-explored research area.

% We present a novel algorithm for multi-conditioned music generation from image and text inputs. Prior works restrict themselves to a single modality (predominantly text) while performing music or general-purpose audio generation. We push the frontier in this space to develop a novel pipeline capable of incorporating more than one input conditioning to produce high-fidelity musical tracks from multi-modal streams. Our proposed \modelname~ leverages a pre-trained stable diffusion model to inject latent image information into our music generation diffusion module, \ourdiffusion~. This enables infusing image conditioning into a text-to-music diffusion model by finetuning only a handful of parameters. To further facilitate such a task, we collect, annotate, and release \ourdataset~ which comprises \ourdatasetsize~ paired $\langle image, text, music \rangle$ samples. To quantitatively measure the synergy between the input image and the generated music sample we also introduce a novel metric \imagemusicmetric~ through contrastive alignment. Finally, our method outperforms existing text-to-music generation approaches on both subjective as well as objective evaluation with a \textbf{relative gain of up to 67.98\%} on FAD$_{\text {VGG}}$ score. We will make the codebase and dataset public upon acceptance. 

\blfootnote{$^*$Equal contribution.}
\blfootnote{$^\dagger$Work done during internship at Adobe Research.}

\end{abstract}

\vspace{-4.9mm}
   
\section{Introduction}
\label{sec:intro}

\begin{figure}
    \centering
    \includegraphics[width=0.48\textwidth]{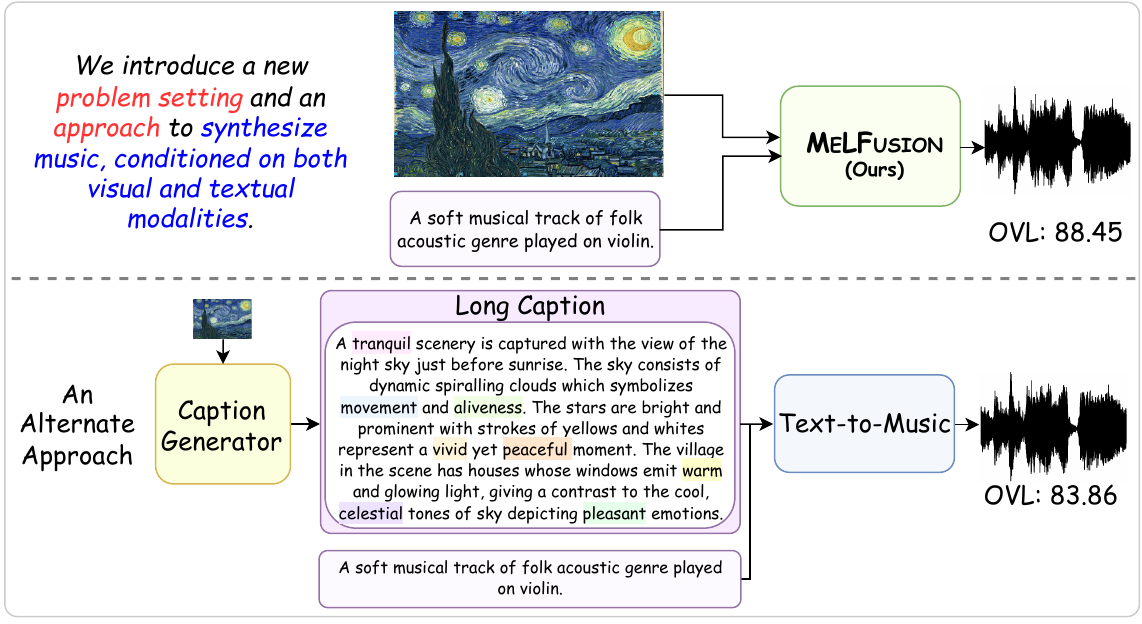}
    \caption{
    We present \modelname, a music diffusion model equipped with a novel ``visual synapse", that can effectively infuse image semantics into a text-to-music diffusion model. This task indeed requires a detailed understanding of the concepts in the image. An alternate approach like using a caption generator to convert image to text space to be further used with existing text-to-music methods leads to a sub-optimal overall audio quality (OVL) score. Our approach can knit together complementary information from both modalities to synthesize high-quality music. 
    % We present a novel approach of multi-conditioned music generation from image-text inputs. We propose \modelname, a novel algorithm that facilitates high-fidelity music generation from multi-modal streams. Prior methods require long, descriptive captions to capture the attributes of the input image. We demonstrate that by infusing image modality along with a short text instruction (e.g., genre and instrument), we achieve substantially high overall audio quality (OVL) scores.
    }
    \label{fig:teaser}
\end{figure}

Music is an essential tool for creative professionals and content creators. It can complement and set the mood for an accompanying still image, animation, video, or even text descriptions while creating a social media post. Finding music that matches a specific setting, can indeed be an arduous task. A conditional music generation approach, that can synthesize a music track by analyzing the visual content and the textual description can find a wide range of practical applications in various fields including social media. 

Inspired by the progress in generative modeling of images, music generation has also garnered significant attention from the community \cite{mousai, melody, musiclm}. 
Recently, \citet{musiclm, musicgen} proposed conditioning in the form of melody or humming.
% - but those act as an auxiliary modality. 
While \citet{ihearyourturecolors} pursue image-guided audio generation. Despite these efforts, music generation conditioned on multiple modalities like text and image, is largely uncharted.
% Efforts thus far have been restricted to generating music (or audio in general) from a single modality (primarily text). 
% These studies notwithstanding, music generation conditioned on more than one independent modality remains largely uncharted. 
% It enhances user engagement for social-media posts, or even convey emotions to targeted audience.

% Given the exponential rise of users across different social media platforms, accompanying musical track generation as a mode of multimedia content creation is a rapidly evolving area of interest. 

Images are more expressive \cite{imageisworththousand} than text-only information and capture more fine-grained semantic information about various visual aspects. 
% For example, a text prompt \textit{`Progression in guitar'} leaves room for ambiguities like which type of guitar is being referred to (acoustic or electric) or how many guitarist(s) are playing the instrument. However, the image associated with this prompt can provide conclusive answers to these queries underpinning the utility of such a multi-modal set up. 
For example, as depicted in Fig \ref{fig:teaser}, to generate a musical track that goes well with a given image, without indeed using it, one has to make the tedious effort of producing long, descriptive captions (either generated by an image captioning model or human annotators) before employing a typical text-to-music generation model. Moreover, the model has to be supplied with critical attributes like \textit{`tranquil'}, \textit{`aliveness'} etc (highlighted in figure) to aptly capture the essence of the image. This poses a major bottleneck in the scalability of such systems especially for social media content creators and necessitates direct image conditioning with textual control in music generation.    

Music is indeed different from generic audio. Music contains an arrangement of elements structured to form a coherent and complete entity. These musical elements include melody, harmony, rhythm, dynamics, and form \cite{schmidt2012basic, temperley2013statistical}. Unlike audio, music contains harmonies from different instruments forming intricate structures. Prior studies show \cite{habibi2014music, wu2013effects, zatorre2003music, norman2019divergence, das2020measurement, fedorenko2012sensitivity} that the human brain is extremely sensitive to disharmony. As a result, the margin of error especially in producing musical pieces is low compared to generic audio tracks. This makes music generation a harder task as the model should be equipped to control the fine-grained nuances of a composition involving melody, the interplay of the instruments, and genre.  
% To this end, it is imperative that an AI-enabled music production model is equipped to control the fine-grained nuances of a composition involving melody, interplay of the instruments, genre, etc.  

% Music generation has been an avenue of interest for artificial intelligence (AI) researchers for a
% considerable amount of time now,
%%%%%%%%%%%%%%%%%%%%%%%%%%%%%%%%%%%%%%%%%%%%%%%%%%
% why adding image modality helps
%%%%%%%%%%%%%%%%%%%%%%%%%%%%%%%%%%%%%%%%%%%%%%%%

% To this end, we present \modelname~ which can incorporate dual conditioning in the form of image and textual descriptions to produce suitable musical tracks. 

An alternative to generating music would be to retrieve them. Retrieval-based systems \cite{mulan, manco2022contrastive} struggle to `match' the right track for a given input prompt thereby limiting their practical applicability in open-world scenarios primarily because (a) they tend to search from a pre-existing collection of tracks and (b) finding the correct association between the input prompt and the audio track can be challenging. 
The problem is inherently complex due to the multifaceted nature of music and the abstract associations between auditory experiences and other sensory modalities. 
% Integrating textual cues involves understanding the semantics and emotions within a given text, while interpreting images demands spatial, acoustic properties, and object recognition, coupled with the extrapolation of associated sentiments and narratives. 

% The recent surge in employing language models (LMs) \cite{flant5, w2vbert, t5} for generative tasks has translated to tremendous success in modeling complex relationships across long-term contexts. A series of works including \cite{musiclm, audiolm} have leveraged language models for audio generation tasks. Another line of work involving diffusion probabilistic models (DPMs) has also gained significant interest \cite{tango, audioldm, melody} in similar tasks including audio and speech generation. Advancements in self-supervised audio representation learning \cite{hubert, tagliasacchi2019self} and audio synthesis \cite{leng2022binauralgrad, revise} furnish the requirements in developing such models. 

% In order to achieve fine-grained control on audio-based modeling, investigations \cite{} suggest to represent the same signal as multiple streams of discrete tokens. 

% Inspired by the recent success of diffusion models in the landscape of generative modeling we propose a novel framework comprising a guided diffusion pipeline to combine textual and visual features into a unified representation to produce high-fidelity music. 

To overcome these shortcomings, we introduce the first music generation model that can be conditioned on image and text instruction. We observe that the features from a pre-trained text-to-image diffusion model that consumes the DDIM-inverted latent of the image can guide a text-to-audio diffusion model. Our key novelty is to facilitate this information exchange by incorporating a ``visual synapse" to the text-to-music model, which includes a set of parameters that learn to combine the signals from both modalities. 

\noindent{\textbf{We summarise our main contributions below:}}
% In summary:
% We hypothesize that a pre-trained text-to-image diffusion model has rich semantic information that can guide an independent text-to-music diffusion model to generate music that is consistent with the image conditioning and textual conditioning.
% Our proposed approach can function with short text instructions and leverages the accompanying image to obtain crucial contextual details. \modelname~ unburdens us from supplementing the model with detailed textual descriptions and can be utilized to produce perceptually superior musical tracks from a handful of instructions such as instrument specification (`violin'), genre (`folk-acoustic'), etc. Our method comprises a pre-trained stable diffusion backbone to obtain the fine-grained features from the decoder layer which it subsequently injects into the cross-attention layers of our text-to-audio diffusion module \ourdiffusion.

% We present a novel task of accompanying music generation for given image and text inputs. To this end, we develop a novel training paradigm  

% \textbf{(1)} To the best of our knowledge, we are the first to propose a novel task of multi-modal music generation where the generation process can be dual-conditioned on both image as well as text prompts.
\textbf{(1)} We formalize a novel task of generating music that is consistent with a reference image and an associated text prompt.

\textbf{(2)} We present \modelname, a novel diffusion model that can address this pragmatic task.  
% generate music conditioned on image and textual signals.
% that enables feature injection between a pre-trained stable diffusion model and a text-to-music diffusion module \ourdiffusion. We impose image conditioning into \ourdiffusion~ by finetuning only a handful of parameters. 

\textbf{(3)} We introduce \ourdataset dataset comprising \ourdatasetsize $\langle \text{image}, \text{text}, \text{music} \rangle$ triplets. To the best of our knowledge, this is the largest collection of these three modalities. Further, we extend the MusicCaps \cite{musiclm} dataset by supplementing the text, and music pairs with suitable images extracted from corresponding YouTube videos or the web. 

% that can facilitate such a task by preparing \ourdataset~ comprising \ourdatasetsize~ samples. The dataset contains paired $\langle image, text, music \rangle$ samples that we collect, extract, and annotate from the web. Furthermore, we extend the MusicCaps \cite{musiclm} dataset by supplementing the text, and music pairs with suitable \textit{images} extracted from the corresponding YouTube videos. 

\textbf{(4)} In order to quantitatively establish the correspondence between the image-music pairs we propose a new metric \imagemusicmetric. We demonstrate that the score follows human perception closely, through a user study.

\textbf{(5)} Finally, our exhaustive experimental results reveal that our approach outperforms existing text-to-music generation pipelines on both subjective as well as objective evaluation with a relative gain of up to \textbf{67.98\%} on FAD score, thereby setting a new benchmark for multi-modal music synthesis.
% on subjective and objective evaluation.

% \input{sec/2_related_works}

\section{Related Works}
\label{sec:related_works}

\noindent{\textbf{Music Generation Approaches:}}
Music generation has garnered significant attention for a considerable amount of time. While some approaches \cite{yang2017midinet, muhamed2021symbolic, dong2018musegan} deploy GANs to tackle this task, \citet{ycart2017study} introduced recurrent neural networks to model polyphonic music. \citet{bassan} proposed an unsupervised segmentation using ensemble temporal prediction errors. Jukebox \cite{dhariwal2020jukebox} tackles the long context of raw audio using a multiscale VQ-VAE to compress it to discrete codes, modeling those using autoregressive Transformers. Another stream of work \cite{gan2020foley, yang2017midinet} that predicts the MIDI notes to produce music has gained popularity in this space. However, the scope of these approaches is relatively limited as they need additional decoders to produce the musical pieces from the notations. 

MusicLM \cite{musiclm} generates high-fidelity music from text descriptions by casting the process of conditional music generation as a hierarchical sequence-to-sequence modeling task. 
% They leverage the audio-embedding network of MuLan \cite{mulan} to extract the representation of the target audio sequence. 
Mubert \cite{mubert} is an API-based service that employs a Transformer backbone. The encoded prompt is used to match the music tags and the one with the highest similarity is used to query the audio generation API. 
% It operates over a relatively smaller set as it produces a combination of audio from a predefined collection. 
MusicGen \cite{musicgen} comprises a single-stage transformer LM together with efficient token interleaving patterns. This eliminates the need for hierarchical upsampling. Despite significant progress, none of these approaches utilize the semantic information of images to condition the audio generation.
% We compare with MusicLM, Mubert and MusicGen in \cref{sec:main_results}.

\noindent{\textbf{Diffusion Models for Music Generation:}}
With the prolific success of diffusion models in conditional image generation, there have been recent efforts in music generation using them. Riffusion \cite{riffusion} base their algorithm on fine-tuning a stable diffusion model \cite{rombach2022high} on mel-spectrograms of music pieces from a paired
music-text dataset. This is one of the first text-to-music generation methods. 
Moûsai \cite{mousai} is a cascading two-stage latent diffusion model that is equipped to produce long-duration high-quality stereo music. 
% It achieves this through employing a specially designed U-Net facilitating high compression rate. 
Noise2Music \cite{noise2music} introduced a series of diffusion models, a generator, and a cascade model. The former generates an intermediate representation
conditioned on text, while the later can produce audio conditioned on the intermediate representation of the text. 
MeLoDy \cite{melody} pursues an LM-guided diffusion model by reducing the forward pass bottleneck and applies a novel dual-path diffusion mode. We find that the visual guidance that is incorporated into our approach significantly enhances the music generation quality when compared to all these approaches. We elaborate this further in \cref{sec:main_results}.

\noindent{\textbf{Diffusion Models for Audio Generation:}}
% Recent years have seen tremendous progress in Denoising Diffusion Probabilistic Models (DDPMs) as a leading approach in generative modeling \cite{tango}.
% In practice, these methods consist of a fixed number of Markov chain steps to convert the original input noise to a meaningful output. 
Diffusion-based methods \cite{huang2022generspeech, huang2022prodiff, huang2022fastdiff, popov2021grad, lam2022bddm, lee2021priorgrad} achieve remarkable results in speech synthesis too. FastDiff \cite{huang2022fastdiff} deploys time-aware location-variable convolutions of diverse receptive field patterns to efficiently model long-term time dependencies with adaptive conditions. AudioLDM \cite{audioldm} is a text-to-audio system that is built on a latent space to learn continuous audio representations from contrastive language-audio pretraining (CLAP) embeddings. 
\citet{tango} simplifies the architecture of AudioLDM, and uses FLAN-T5 \cite{flant5} as the text encoder.
% The CLAP model helps in training the latent diffusion model with audio embeddings while providing text embeddings as the condition during sampling. 
% The success of these approaches notwithstanding, they are known to have high latency pertaining to the iterative generation process in a high-dimensional space. A plausible workaround, as shown by \cite{vahdat2021score, rombach2022high} is to leverage them in a relatively smaller latent space. 
Another line of work \cite{yang2023diffsound, garcia2023vampnet} involves text-conditional discrete diffusion models to generate discrete tokens as a representation for spectrograms. However, the quality of the sound produced by such methods leaves room for improvements in terms of both subjective and objective qualities, thereby limiting their practical usability. 
In contrast to these approaches, our method generates music samples conditioned on visual and textual signals. 

\section{Synthesizing Music from Image and Text}
\label{sec:method}

\begin{figure*}
    \centering
    \includegraphics[width=\textwidth]{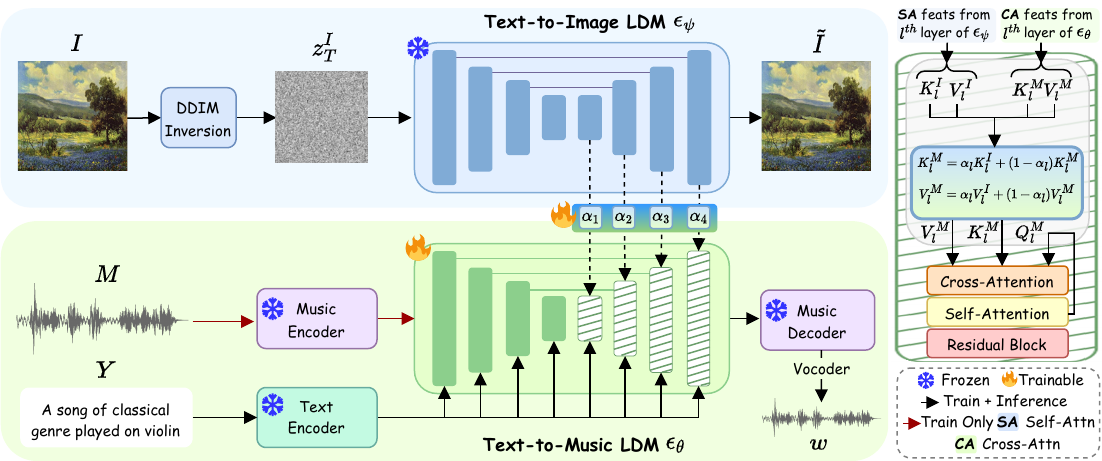}
    \caption{Our approach \textbf{\modelname} generates music waveform $\bm{w}$ conditioned on an image $\bm{I}$ and a given textual instruction $\bm{Y}$. Visual semantics from  $\bm{I}$ is instilled into a text-to-music diffusion model (bottom green box) using a pre-trained and frozen text-to-image diffusion model (top blue box). The image $\bm{I}$ is first DDIM inverted into a noisy latent $\bm{z}^I_T$. The self-attention features from the decoder layers of the text-to-image LDM that consumes $\bm{z}^I_T$ is infused into the cross-attention features of text-to-music LDM decoder layers, modulated by learned $\alpha$ parameters. 
    This fusion operation that happens in the decoder (green stripes) is detailed on the right side of the figure. The music encoder projects the spectrogram representation of the music to the latent space, and the music decoder retrieves back the spectrograms. Finally, a vocoder generates the waveform $\bm{w}$ from the spectrograms. Please refer to \cref{sec:method} for more details. 
    % Overview of \textbf{\modelname}. Our method comprises a pre-trained stable diffusion backbone that takes in the DDIM inverted image to obtain the reconstruction features from the decoder layer which it subsequently injects into the cross-attention layers of our \ourdiffusion~ (text-to-audio diffusion module). The internals of the decoder layers of the \ourdiffusion~ (green stripes) UNet is further elaborated on the right side. The audio encoder input (present during training only) is used to train the diffusion model. A Flan-T5 model acts as a text encoder. The output of \ourdiffusion is passed through Audio decoder and HiFi GAN Vocoder \cite{hifigan} to generate the audio output
    }
    \label{fig:main-figure}
\end{figure*}

We propose to learn a conditional distribution $\mathcal{M}(\bm{w}|\bm{I}, \bm{Y})$, that can generate music waveforms $\bm{w}$ from an image $\bm{I}$ and a paired textual description $\bm{Y}$. 
We materialize $\mathcal{M}$ as \modelname, a diffusion model that can succinctly interleave the semantic cues from the image and textual modality while generating acoustically pleasing music. 

\cref{fig:main-figure} provides an overview of our approach.
On a high level, our novel methodology consists of two sub-components: 1) an approach to extract relevant visual information from the image conditioning $\bm{I}$ and 2) a method to induce this conditioning into the text-to-music generative model, in a parameter efficient way. We describe each of these in the subsequent subsections.
% \subsection{Problem Formulation}

% \subsection{Recap of Text-to-Image LDM}

\subsection{Extracting Visual Guidance} \label{sec:extracting_guidance}
Latent diffusion models (LDMs) for text-to-image generation \cite{rombach2022high} have had phenomenal success in generating high-quality images that are well-grounded in their textual conditioning. We hypothesize that the latent representations and their transformations encode rich semantic knowledge, that can guide our audio diffusion model. In our exploration, we make use of a pre-trained Stable Diffusion model \cite{rombach2022high}. It contains a VQ-VAE \cite{van2017neural} for encoding and decoding the image to the latent space, a text encoder, and a UNet \cite{unet} that carries out the diffusion process on the latent. The UNet contains an encoder, a bottleneck layer, and a decoder. Each encoder and the decoder further contain a set of blocks with cross-attention layers, self-attention layers, and convolutional layers. Given any intermediate latent image feature $\bm{f} \in \mathbb{R}^{(w \times h) \times d}$, a single self-attention \cite{attention} operation consist of $\bm{Q} = \bm{W}^q \bm{f}$, $\bm{K} = \bm{W}^k \bm{f}$, $\bm{V} = \bm{W}^v \bm{f}$:
\begin{equation}
    \text{Attention}(\bm{Q}, \bm{K}, \bm{V}) = \text{Softmax}\left(\frac{\bm{Q} \bm{K}^T}{\sqrt{d_k}}\right)\bm{V},
    \label{eqn:attn}
\end{equation}
where $d_k$ is the dimension of the query and key features. During cross-attention, the key and value matrices operate on the external text conditioning $\bm{c} \in \mathbb{R}^{s \times d_k}$: $\bm{K} = \bm{W}^k \bm{c}$, $\bm{V} = \bm{W}^v \bm{c}$. Here, $\bm{W}^q, \bm{W}^k $ and $ \bm{W}^v$ are the attention weight matrices that transform either the image features or text conditions into the output of each block.

We want to transfer over the semantic information that is present within these attention layers corresponding to the image $\bm{I}$ into the music LDM. For this, we first invert $\bm{I}$ into the latent space using DDIM Inversion \cite{song2020denoising} to get $\bm{z}^{I}_{T}$. This will guarantee that we will be able to generate $\bm{I}$ from $\bm{z}^{I}_{T}$. Next, we do the reverse diffusion steps using a pre-trained text-to-image LDM starting from $\bm{z}^{I}_{T}$ and save the \textit{self-attention features} $\bm{K} = \bm{W}^k \bm{f}$, $\bm{V} = \bm{W}^v \bm{f}$, to be injected into the music LDM. The intuition behind leveraging the self-attention features is that they control the feature transformations responsible for generating the visual semantics of the image. This is mathematically evident from \cref{eqn:attn}. 
% We elaborate how these features are used in the text-to-music diffusion model next.
In the subsequent section, we elaborate on how we construct the ``synapse" that can transfer the guidance information from $\bm{I}$ to the music-diffusion model.

\subsection{Text-to-Music LDM with Visual Synapse}\label{sec:visual_synapse}
Inspired by recent text-to-audio \cite{audioldm,tango} generation approaches, our text-to-music model is also formulated as a latent diffusion model. During training, the music waveform $\bm{w}$ is first converted to a spectrogram $\bm{s} \in 
\mathbb{R}^{E \times F}$, which is a visual representation obtained via Fourier Transformation on $\bm{w}$. $E$ and $F$ denote the number of time slots and frequency slots respectively. Then we encode $\bm{S}$ using Audio-VAE \cite{audioldm} to get a latent representation $\bm{z}_1^M \in \mathbb{R}^{C \times E/r \times F/r}$, where $C$ is the number of channels and $r$ is the compression level.

The forward diffusion process involves corrupting $\bm{z}_1^M$ using a Markovian noise process $q$, which gradually adds noise to $\bm{z}_1^M$ through $\bm{z}_T^M$ over $T$ steps with the following Gaussian function:

\vspace{-0.11in}

\begin{equation}
    q(\bm{z}_{t}^M | \bm{z}_{t-1}^M) = \mathcal{N}(\bm{z}_t^M; \sqrt{1 - \beta_t}\bm{z}_{t-1}^M, \beta_t 
\textbf{I}),
\end{equation}
where $\beta_t$ is a predetermined variance schedule. This iterative sampling process can be approximated by a deterministic non-Markovian process as follows \cite{song2020denoising}:
\begin{align} 
q(\bm{z}_{t}^{M} | \bm{z}_{1}^{M}) &=  \mathcal{N}(\bm{z}_{t}^{M}; \sqrt{\bar{\gamma}_t}\bm{z}_{1}^{M}, (1 - \bar{\gamma}_t) \textbf{I}) \\ 
 &=  \sqrt{\bar{\gamma}_t}\bm{z}_{1}^{M} + \epsilon \sqrt{(1 - \bar{\gamma}_t)}, \epsilon \sim \mathcal{N}(\boldsymbol{0}, \textbf{I}) \label{eqn:noise}
\end{align}
where $\gamma_t = 1 - \beta_t$ and $\bar{\gamma}_t = \prod_{r=0}^t \gamma_r$.

In the reverse diffusion process, an LDM  $\epsilon_\theta(\cdot,\cdot,\cdot)$ (implemented as a UNet), learns to de-noise $\bm{z}_T^M \sim \mathcal{N}(\boldsymbol{0}, \textbf{I})$ to recover $\bm{z}_1^M$. The architecture of the UNet is kept exactly similar to the text-to-image UNet described in \cref{sec:extracting_guidance}. To incorporate the additional guidance from image conditioning the \textit{cross-attention} key and value features $\bm{K}_l^{M}$ and $\bm{V}_l^{M}$ in each of the decoder layer $l$ of the UNet is modified as follows:

\vspace{-0.18in}

\begin{align} 
\bm{K}^M_l &=  \alpha_l \bm{K}^I_l + (1 - \alpha_l) \bm{K}^M_l \\ 
\bm{V}^M_l &=  \alpha_l \bm{V}^I_l + (1 - \alpha_l) \bm{V}^M_l\label{eqn:synapse},
\end{align}
where $\bm{K}^I_l$ and $\bm{V}^I_l$ are the \textit{self-attention} features for the corresponding layer $l$ of the image conditioning LDM from \cref{sec:extracting_guidance}. Most importantly, the convex combination between these features is modulated by \textit{learned layer specific $\alpha$ parameters}. We find that this simple formulation elegantly incorporates the image guidance into the text-to-music diffusion model without hampering its expressivity. As the $\alpha$ parameters facilitate the information exchange between the text-to-audio and text-to-image diffusion models, analogous to how a synapse in a nervous system facilitates the transfer of electrical and chemical signals between neurons, we refer to this handshake as the \textit{visual synapse of a text-to-music LDM}. 

Finally, the parameters of the LDM $\theta$ and the $\alpha$ parameters are trained end-to-end with the following loss function: 

\vspace{-0.17in}

\begin{equation}
    \mathcal{L} = \mathbb{E}_{t\sim[1,T], \bm{z}_1^M, \bm{\epsilon}^M_t \sim \mathcal{N}(\boldsymbol{0}, \textbf{I})}\lVert \bm{\epsilon}^M_t - \epsilon_\theta(\bm{z}_t^M, \bm{c}, t) \rVert^2
    \label{eqn:loss}
\end{equation}

\subsection{Overall Framework}

\begin{algorithm}[!t]
\small
\caption{\modelname: Training}
\label{algo:training}
\begin{algorithmic}[1]
\Require{Image: $\bm{I}$; Text: $\bm{Y}$; Music: $\bm{M}$; 
Pre-trained Text-to-Image LDM: $\epsilon_\psi(\cdot,\cdot,\cdot)$; Image Encoder: $\mathcal{E}^I(\cdot)$; Music Encoder: $\mathcal{E}^M(\cdot)$; Text Encoder: $\mathcal{T}^M(\cdot)$; Text-to-Music LDM: $\epsilon_\theta(\cdot,\cdot,\cdot)$; Number of Diffusion Steps: $T$.
% Learnable mixing coefficient for layer $l$: $\alpha_l$.
}
\Ensure{Trained Text-to-Music LDM: $\epsilon_\theta(\cdot,\cdot,\cdot)$, Learned mixing coefficient $\alpha$, for each decoder layer $l$ of LDM: $\{\alpha_l\}$.}
\State $\bm{z}^{I}_{T} \leftarrow \text{DDIM\_Invert}(\mathcal{E}^I(\bm{I}))$ \Comment{\textit{Initialize Image Latent.}}
\State $\{\bm{\epsilon}_1^M, \cdots \bm{\epsilon}_T^M\} \leftarrow \text{Forward\_Diffusion}(\mathcal{E}^M(\bm{M}))$ \Comment{\textit{Targets.}}
\State $\bm{z}^{M}_{T} \sim \mathcal{N}(\bm{0}, \textbf{I})$ \Comment{\textit{Initialize Music Latent.}}
\State $\bm{c} \leftarrow \mathcal{T}^M(\bm{Y})$ \Comment{\textit{Encoding Text.}}
\For{$t \in \{T, \cdots, 1\}$} \Comment{\textit{For each denoising step.}}
\For{each layer $l$ in decoder of LDM} 
\State $\bm{K}^I_l$, $\bm{V}^I_l \leftarrow$ Self-attention features of $\epsilon_\psi(\bm{z}^{I}_{t},\emptyset, t)$.
\State $\bm{K}^M_l,\bm{V}^M_l\leftarrow$Cross-attention features of $\epsilon_\theta(\bm{z}^{M}_{t},\bm{c},t)$.
\State $\bm{K}^M_l \leftarrow \alpha_l \bm{K}^I_l + (1 - \alpha_l) \bm{K}^M_l$  \Comment{\textit{Key update.}}
\State $\bm{V}^M_l \leftarrow \alpha_l \bm{V}^I_l + (1 - \alpha_l) \bm{V}^M_l$  \Comment{\textit{Value update.}}
\EndFor
\State $\mathcal{L} \leftarrow ||\bm{\epsilon}_t^M - \epsilon_\theta(\bm{z}^{M}_{t}, \bm{c}, t)||^2$ \Comment{\textit{\cref{eqn:loss}}}
\State Optimize $\theta$ and all $\alpha$ parameters to reduce $\mathcal{L}$.
\EndFor
\State \Return $\epsilon_\theta(\cdot,\cdot,\cdot)$, $\{\alpha_l\}$.
\end{algorithmic}
\end{algorithm}
We summarize the overall flow of \modelname~during training in \cref{algo:training}. Our key novelty is to introduce a channel through which we can guide the text-to-music diffusion model toward the semantic concepts contained in the corresponding image conditioning. This ``synapse" is detailed in Line 7 to Line 10. The rest of the algorithm follows the standard LDM training flow.

During inference, we make use of the trained text-to-image and text-to-music diffusion models, along with the learned $\alpha$ parameters. As seen in Lines 8 and 9 in \cref{algo:sampling}, the cross-attention features of the text-to-music LDM decoder are updated to incorporate the visual conditioning in each denoising step. Once the denoising (Line 10) is complete, the latent representation is projected back into a spectrogram using the decoder of Audio VAE \cite{audioldm}, and then the waveform is generated using HiFi-GAN vocoder \cite{hifigan} in Lines 11 and 12 respectively.

\begin{algorithm}[!t]
\small
\caption{\modelname: Sampling}
\label{algo:sampling}
\begin{algorithmic}[1]
\Require{Image: $\bm{I}$; Text: $\bm{Y}$;
Pre-trained Text-to-Image LDM: $\epsilon_\psi(\cdot,\cdot,\cdot)$; Image Encoder: $\mathcal{E}^I(\cdot)$; Text Encoder: $\mathcal{T}^M(\cdot)$; Trained Text-to-Music LDM: $\epsilon_\theta(\cdot,\cdot,\cdot)$; Learned mixing coefficient $\alpha$, for each decoder layer $l$ of LDM: $\{\alpha_l\}$; Number of Diffusion Steps: $T$; Music Decoder: $\mathcal{D}^M(\cdot)$; Vocoder $\mathcal{V}(\cdot)$.
}
\Ensure{Music Waveform: $\bm{w}$}
\State $\bm{z}^{I}_{T} \leftarrow \text{DDIM\_Invert}(\mathcal{E}^I(\bm{I}))$ \Comment{\textit{Initialize Image Latent.}}
\State $\bm{z}^{M}_{T} \sim \mathcal{N}(\bm{0}, \textbf{I})$ \Comment{\textit{Initialize Music Latent.}}
\State $\bm{c} \leftarrow \mathcal{T}^M(\bm{Y})$ \Comment{\textit{Encoding Text.}}
\For{$t \in \{T, \cdots, 1\}$} \Comment{\textit{For each denoising step.}}
\For{each layer $l$ in decoder of LDM} 
\State $\bm{K}^I_l$, $\bm{V}^I_l \leftarrow$ Self-attention features of $\epsilon_\psi(\bm{z}^{I}_{t},\emptyset, t)$.
\State $\bm{K}^M_l,\bm{V}^M_l\leftarrow$Cross-attention features of $\epsilon_\theta(\bm{z}^{M}_{t},\bm{c},t)$.
\State $\bm{K}^M_l \leftarrow \alpha_l \bm{K}^I_l + (1 - \alpha_l) \bm{K}^M_l$  \Comment{\textit{Key update.}}
\State $\bm{V}^M_l \leftarrow \alpha_l \bm{V}^I_l + (1 - \alpha_l) \bm{V}^M_l$  \Comment{\textit{Value update.}}
\EndFor
\State $\bm{z}^{M}_{t} \leftarrow \bm{z}^{M}_{t} - \epsilon_\theta(\bm{z}^{M}_{t},\bm{c},t)$ \Comment{\textit{Reverse Diffusion Step.}}
\EndFor
\State $\bm{s} \leftarrow \mathcal{D}^M(\bm{z}^{M}_{0})$ \Comment{\textit{Generate Spectrograms.}}
\State $\bm{w} \leftarrow \mathcal{V}(\bm{s})$ \Comment{\textit{Generate Waveform from Spectrograms.}}
\State \Return $\bm{w}$.
\end{algorithmic}
\end{algorithm}

\section{Experiments and Results}
\label{experiments_and_results}
To complement our newly introduced problem setting which generates music conditioned on visual and textual modality, we introduce \underline{a new dataset}, a \underline{new evaluation metric}, and come up with a \underline{strong baseline} by extending a state-of-the-art text-to-music method to consume image modality. We explain each of these in the subsequent sections.

\subsection{Datasets}
\label{datasets}

% \begin{table}
% \centering
% \resizebox{\columnwidth}{!}
% {\begin{tabular}{c c c}
% \toprule
% \textbf{Genre} & \textbf{\# Pieces} & \textbf{Percentage (\%) in Dataset}\\
% \midrule
% Rock & 689 & 6.01 \\
% Classical & 711 & 6.02 \\
% Latin & 732 & 6.04 \\
% Hip-Hop & 767 & 6.70 \\
% Folk Acoustic & 671 & 5.85 \\
% \bottomrule
% \end{tabular}}
% \caption{\ourdataset~ covers musical pieces from 15 different genres including Rock, Classical, Latin, etc. Each genre is divided into 22 subcategories. The distribution of randomly chosen 5 categories is presented here.}
% \label{tab:dataset}
% \end{table}

To the best of our knowledge, there is no publicly available dataset that contains the $\langle \text{Image}, \text{Text}, \text{Music} \rangle$ triplets that are required to train and evaluate \modelname. We collect a new dataset \ourdataset, which contains \ourdatasetsize manually annotated triplets of $\langle \text{Image}, \text{Text}, \text{Music} \rangle$. Further, we extend the MusicCaps \cite{musiclm} dataset which contains $\langle \text{Text}, \text{Music} \rangle$ pairs by adding the corresponding image. 

% As there is no pre-existing dataset that contains $\langle Image, Text, Audio \rangle$ pairs, to facilitate research in this direction we extend the MusicCaps \cite{musiclm} dataset and also collect, prepare, and release a new dataset \ourdataset. Since we are introducing an additional modality by supplementing the dataset with images, we also perform analysis on the quality of the collected samples (details in section \ref{cimp_metric} during data curation. This ensures that the collected images are perceptually relevant to the corresponding music samples.     

\begin{figure}
    \centering
    \includegraphics[width=0.4\textwidth]{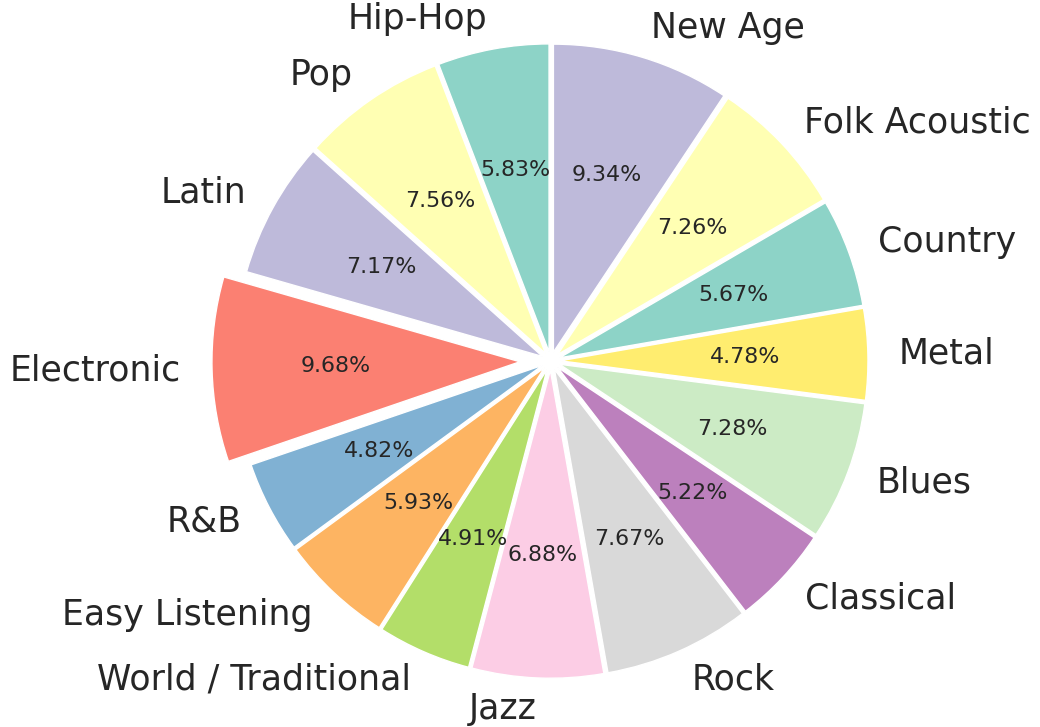}
    \caption{The distribution of different genres in \ourdataset.}
    \label{fig:dataset_pie_chart}
\end{figure}

\begin{figure}
    \centering
    \includegraphics[width=\columnwidth]{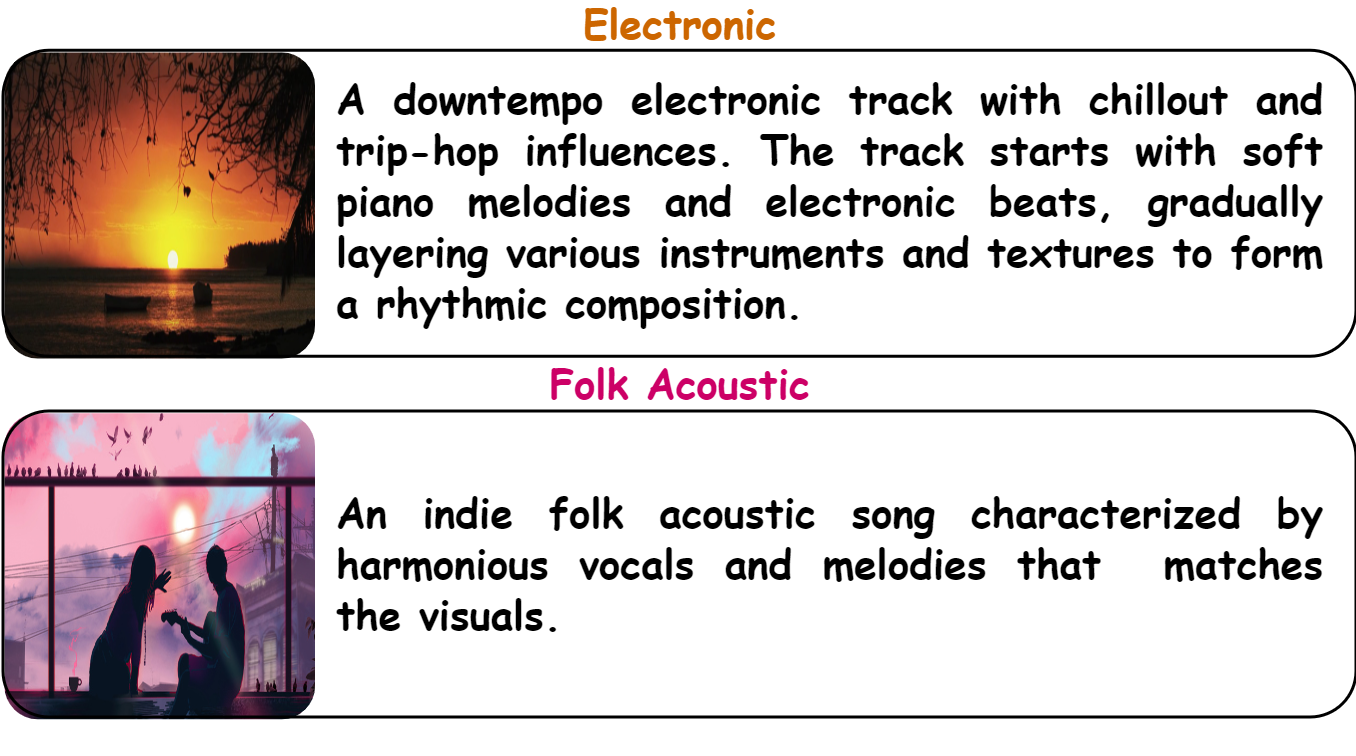}
    \caption{Some image and text pairs from \ourdataset. We include more examples in the Appendix.}
    \label{fig:dataset_main_paper}
\end{figure}

\noindent{\textbf{\ourdataset:}}
We hired $18$ professional annotators to find $10$-second snippets of YouTube videos corresponding to $15$ predefined genres. The annotators are trained musicians with at least 5 years of practice. For each of these videos, they were asked to provide (a) a free-form text description for up to three sentences, expressing the composition and (b) any other music-related details such as describing the genre, mood, tempo, singer voices, instrumentation, dissonances, rhythm, etc. A carefully selected frame and music from the snippet along with text description from annotators forms $\langle \text{Image}, \text{Text}, \text{Music} \rangle$ triplets. We perform strict sanity checks to ensure the quality of these triplets in \ourdataset. \cref{fig:dataset_main_paper} shows some image and text samples from the dataset and \cref{fig:dataset_pie_chart} shows the distribution of different genres in \ourdataset.  Before annotating YouTube snippets (containing music-albums, art-performance, ensembles etc.), they were asked to check for complementary relevance between visuals and music. Further, we perform manual validation to filter lower-quality samples. We include more examples and more statistics of the dataset in the Appendix.


\noindent{\textbf{Extended MusicCaps:}}
MusicCaps \cite{musiclm} is a subset of the AudioSet \cite{gemmeke2017audio} dataset, which contains music and a corresponding textual description of the same. We carefully choose two images from the web or YouTube that can go well with each datapoint in MusicCaps, thereby extending $\langle \text{Text}, \text{Music} \rangle$ pairs to $\langle \text{Image}, \text{Text}, \text{Music} \rangle$ triplets. We defer more details to the Appendix.

% This is a music caption dataset comprising music clips from AudioSet \cite{gemmeke2017audio} paired with corresponding text descriptions in English. The collection consists of a total of 5,521 examples, out of which 2,858 are from the AudioSet eval and 2,663 are from the AudioSet train split. The authors further tag 1,000 samples as a balanced subset of the dataset - equally divided across genres. All examples in the balanced subset are from the AudioSet eval split. As our setup is not restricted to text and requires joint conditioning in the form of images as well, we supplement this dataset by collecting 2 carefully chosen image frames for each of the 10-second samples from the corresponding YouTube video or web. Due to the unavailability of some links, we were able to collect a total of \musiccapssamples~ samples which we divided into a 60\%/20\%/20\% split for train/validation/test respectively. 

\begin{table*}
\centering
\resizebox{\textwidth}{!}
{\begin{tabular}{l|cc|cccc|cc|cccc|cc}
\toprule
\multirow{3}{*}{ \bf Model } & \multirow{3}{0.7cm}{ \bf Txt } & \multirow{3}{0.7cm}{ \bf Img} & \multicolumn{6}{c|}{ \bf MusicCaps } & \multicolumn{6}{c}{ \bf \ourdataset } \\
\cmidrule { 4 - 15 }
 &  &  & \multicolumn{4}{c|}{ \bf Objective metrics } & \multicolumn{2}{c|}{ \bf Subjective metrics } & \multicolumn{4}{c|}{ \bf Objective metrics } & \multicolumn{2}{c}{ \bf Subjective metrics }\\
% \cmidrule { 4 - 13 }
& & & \textbf{FAD}$\downarrow$ & \textbf{KL}$\downarrow$ & \textbf{FD}$\downarrow$ & \textbf{\imagemusicmetric}$\uparrow$ & \textbf{OVL}$\uparrow$ & \textbf{REL}$\uparrow$ & \textbf{FAD}$\downarrow$ & \textbf{KL}$\downarrow$ & \textbf{FD}$\downarrow$ & 
\textbf{\imagemusicmetric}$\uparrow$ & \textbf{OVL}$\uparrow$ & \textbf{REL}$\uparrow$ \\
\midrule
Riffusion \cite{riffusion} & \textcolor{ForestGreen}{\ding{51}} & \textcolor{OrangeRed}{\ding{55}} & 13.40 & 1.19 & - & - & 79.48 & 75.60 & 14.06 & 1.42 &  32.64 & - & 80.11 & 76.26 \\
MuBERT \cite{mubert} & \textcolor{ForestGreen}{\ding{51}} & \textcolor{OrangeRed}{\ding{55}} & 9.60 & 1.58 & - & - & 77.59 & 77.93 & - & - & - & - & - & - \\
MusicLM \cite{musiclm} & \textcolor{ForestGreen}{\ding{51}} & \textcolor{OrangeRed}{\ding{55}} & 4.00 & 1.01 & - & - & 81.51 & 82.65 & 3.62 & 0.93 & 23.44 & - & 83.86 & 84.27 \\
Moûsai \cite{mousai} & \textcolor{ForestGreen}{\ding{51}} & \textcolor{OrangeRed}{\ding{55}} & 7.50 & 1.59 & - & - & 75.94 & 77.33 & 9.13 & 1.63 & 31.51 & - & 75.11 & 74.32 \\
% TANGO $^*$ & \ding{51} & \ding{55} & 23.31 & 1.26 & 1.59 & & \\
Noise2Music \cite{huang2023noise2music} & \textcolor{ForestGreen}{\ding{51}} & \textcolor{OrangeRed}{\ding{55}} & 2.13 & - & - & - & 81.13 & 79.88 & - & - & - & - & - & - \\
MeLoDy \cite{melody} & \textcolor{ForestGreen}{\ding{51}} & \textcolor{OrangeRed}{\ding{55}} & 5.41 & - & - & - & 80.61 & 79.25 & - & - & - & - & - & - \\
MusicGen \cite{musicgen} & \textcolor{ForestGreen}{\ding{51}} & \textcolor{OrangeRed}{\ding{55}} & 3.40 & 1.23 & - & - & 83.57 & 83.18 & 3.28 & 1.21 & 23.60 & - & 84.61 & 83.25 \\ % [-1.5ex]
% \hdashline\\[-1.5ex]
% \customdottedline
\midrule
% Baseline (\ourdiffusion + ITC)
MusicLM + InstructBLIP \cite{instructBLIP} & \textcolor{ForestGreen}{\ding{51}} & \textcolor{ForestGreen}{\ding{51}} & 4.12 & 1.18 & 25.68 & 0.55 & 80.21 & 79.85 & 3.88 & 1.07 & 24.96 & 0.63 & 81.18 & 82.42 \\
TANGO++& \textcolor{ForestGreen}{\ding{51}} & \textcolor{ForestGreen}{\ding{51}} & 3.05 & 1.17 & 23.91 & 0.68 & 84.62 & 83.96 & 2.93 & 1.14 & 22.16 & 0.71 & 85.52 & 84.81 \\
\CC{}\textbf{\modelname\ (Ours)} & \CC{}\textcolor{ForestGreen}{\ding{51}} & \CC{}\textcolor{ForestGreen}{\ding{51}} & \CC{}\textbf{1.12} & \CC{}\textbf{0.89} & \CC{}\textbf{22.65} & \CC{}\textbf{0.76} & \CC{}\textbf{86.78} & \CC{}\textbf{85.92} & \CC{}\textbf{1.05} & \CC{}\textbf{0.72} & \CC{}\textbf{20.49} & \CC{}\textbf{0.83} & \CC{}\textbf{88.45} & \CC{}\textbf{87.39} \\%[-1.3ex]
% \customdottedline %[-3.0ex]
\midrule
% $\Delta_{\text{\modelname - MusicGen}}$ & - & - & \colorbox{increase}{+2.28} & \colorbox{increase}{+0.34} & - & \colorbox{increase}{+3.21} & \colorbox{increase}{+2.74} & \colorbox{increase}{+2.23} & \colorbox{increase}{+0.49} & \colorbox{increase}{+3.11} & \colorbox{increase}{+3.84} & \colorbox{increase}{+4.14} \\
\textcolor{blue}{$\Delta_{\text{\modelname - MusicGen}}$} & - & - & \colorbox{increase}{+67.05\%} & \colorbox{increase}{+27.64\%} & - & - & \colorbox{increase}{+3.84\%} & \colorbox{increase}{+3.29\%} & \colorbox{increase}{+67.98\%} & \colorbox{increase}{+40.49\%} & \colorbox{increase}{+13.17\%} & - & \colorbox{increase}{+4.53\%} & \colorbox{increase}{+4.97\%} \\
\bottomrule
\end{tabular}}
\caption{
Our proposed approach \modelname offers significant gains over state-of-the-art text-to-music methods (first section), and our adapted text-and-image conditioned baselines (second section) across multiple objective and subjective metrics on two datasets. \imagemusicmetric is applicable only when the model is conditioned on visual modality. We skip comparison with MuBERT, Noise2Music, and MeLoDy on \ourdataset dataset as their codebases are not public. Please refer to \cref{sec:main_results} for more details. 
% Comparison of \modelname~ with various baselines on music track generation task when evaluated on the (Extended) MusicCaps dataset and \ourdataset. The results for the baseline methods are obtained by training the model from the implementation by the authors. For unreported cases, the codebase were not available publicly
}
\label{tab:main_table}
\end{table*}

% \begin{table}
%     \centering
%     \begin{tabular}{cccccccc}
%          &  &  &  &  &  &  & \\
%          &  &  &  &  &  &  & \\
%          &  &  &  &  &  &  & \\
%          &  &  &  &  &  &  & \\
%          &  &  &  &  &  &  & \\
%     \end{tabular}
%     \caption{Caption}
%     \label{tab:my_label}
% \end{table}

\subsection{Evaluation Metrics}
\label{evaluation_metrics}
We use objective evaluation and human subjective evaluation metrics to measure the efficacy of \modelname. 

\customsubsubsection{Objective Evaluation:}
% \noindent{\textbf{Objective Evaluation:}}
Following previous works \cite{tango, audioldm, audiogen}, Fréchet Audio Distance
(FAD),  Fréchet Distance (FD) and KL Divergence scores are used for objective evaluation. FAD \cite{audiogen} is a perceptual metric that is adapted from Fréchet Inception Distance
(FID) for the audio domain. It uses a VGG-like backbone \cite{hershey2017cnn} for feature extraction. FD is similar to FAD but uses PANNs \cite{kong2020panns} as the feature extractor.
Unlike reference-based metrics, FAD and FD measure the distance between the
generated audio distribution and the real audio distribution without using any reference audio samples. On the other hand, KL Divergence \cite{audiogen} is a reference-dependent metric that computes the divergence between
the distributions of the original and generated audio samples based on the labels generated by a pre-trained
classifier. While FAD is more related to human perception, KL Divergence captures the similarities
between the original and generated audio signals based on broad concepts present in them. 

FAD, FD, and KL Divergence score captures the `goodness' of generated music, while it doesn't measure whether the generated music is consistent with the image conditioning. We identify this as a gap and propose \imagemusicmetric metric.

% Objective Evaluation In this work, we used two commonly used objective metrics: Fréchet Audio Distance
% (FAD) and KL divergence. FAD \cite{audiogen} is a perceptual metric that is adapted from Fréchet Inception Distance
% (FID) for the audio domain. Unlike reference-based metrics, it measures the distance between the
% generated audio distribution and the real audio distribution without using any reference audio samples. On the other hand, KL divergence \cite{audiogen} is a reference-dependent metric that computes the divergence between
% the distributions of the original and generated audio samples based on the labels generated by a pretrained
% classifier. While FAD is more related to human perception, KL divergence captures the similarities
% between the original and generated audio signals based on broad concepts present in them.

\noindent{\textbf{Image Music Similarity Metric (\imagemusicmetric):}}
\label{cimp_metric}
CLIP score is one of the widely used metrics for measuring the similarity between an image and a corresponding textual description. 
$N$ pairs of images and texts are passed through respective encoders (pre-trained using CLIP loss \cite{clip}) to obtain corresponding feature embeddings which are used to compute CLIP score matrix $\mathcal{A}_{\text{CLIP}} \in \mathbb{R}^{N \times N}$. In a very similar fashion, CLAP scores are computed amongst $N$ audio-text pairs yielding CLAP score matrix $\mathcal{A}_{\text{CLAP}} \in \mathbb{R}^{N \times N}$ \cite{clap}. It is worth noting that in both the matrices the columns represent text modality. This motivates us to develop a metric \imagemusicmetric, which is a measure of the perceptual
similarity between given image-music pairs bridged by the text modality. In particular, we use CLIP image and text encoders which are contrastively aligned \cite{clip} to compute the image and text feature embeddings. As a second step, we leverage language as the bridging modality by freezing the CLIP text encoder and aligning the music (audio) encoder via contrastive training \cite{clap}. Finally, for $\langle \text{Image}, \text{Text}, \text{Music} \rangle$ pairs we obtain \imagemusicmetric by suitably combining $\mathcal{A}_{\text{CLIP}}$ and $\mathcal{A}_{\text{CLAP}}$ using the given mathematical expression:

\begin{equation}
\mathcal{A}_{\text{\imagemusicmetric}} = \mathcal{A}_{\text{CLIP}}\;\mathcal{A}_{\text{CLAP}}^{T}
\end{equation}
% CLIP score is one of the widely used metrics for quality assessment of generated images using textual prompts. $N$ pairs of images and lingustic expressions are passed through respective encoders (pretrained using CLIP loss \cite{clip}) to obtain corresponding feature embeddings which are used to compute CLIP score matrix $\mathcal{A}_{\text{CLIP}} \in \mathrm{R}^{N \times N}$. In a very similar fashion, CLAP scores are computed amongst $N$ audio-text pairs yielding CLAP score matrix $\mathcal{A}_{\text{CLAP}} \in \mathrm{R}^{N \times N}$ \cite{clap}. It is worth noting that in both the matrices the columns represents text modality. Taking inspiration from the above, we develop a metric \imagemusicmetric which is a measure of the perceptual
% similarity between given image-music pairs bridged by the text modality. In particular, we use CLIP image and text encoders which are contrastively aligned \cite{clip} to compute the image and text feature embeddings. As a second step, we leverage language as the bridging modality by freezing the CLIP text encoder and aligning the music (audio) encoder via contrastive training \cite{clap}. Finally, for $\langle$ Image, Text, Music $\rangle$ pairs we obtain \imagemusicmetric by suitably combining $\mathcal{A}_{\text{CLIP}}$ and $\mathcal{A}_{\text{CLAP}}$ using the given mathematical expression:

% \begin{equation}
%     \mathcal{A}_{\text{CIMP}} = \mathcal{A}_{\text{CLIP}}\;\mathcal{A}_{\text{CLAP}}^{T}
% \end{equation}

\customsubsubsection{Subjective Evaluation:}
% \noindent{\textbf{Subjective Evaluation:}}
Following earlier works in text-to-audio generation \cite{tango, audioldm, audiogen}, we use overall audio quality (OVL) and relevance to image-text inputs (REL) to analyze the results of our subjective user study involving 75 participants. They were presented with 100 randomly generated samples from \modelname. Each of the metrics (OVL and REL) is a score between $1$-$100$ with $1$ being the lowest. For the OVL score, the users are asked to assign a score based on how perceptually realistic the generated audio is, while the REL score requires them to carefully examine the image and the text prompts before providing a rating based on their relevance with the generated music. We add more details of the user study in the Appendix.

% Inspired by \cite{tango, audioldm, audiogen} we use the overall audio quality (OVL) and relevance to image-text inputs (REL) as our subjective metrics. Both these metrics denote a score in a scale of $1-100$ with $1$ being the lowest. For the OVL score the users were asked to assign a score based on how perceptually realistic the generated audio is while the REL score required them to carefully examine both the image and the text prompts before assigning a score.  

% \noindent{\textit{Subject matter expert evaluation}} Taking one step further, we studied the genre wise music generation capabilities    

\subsection{Baseline Methods}

We compare \modelname against strong baselines to test its mettle. To the best of our knowledge, there doesn't exist a music diffusion model that is conditioned on visual and textual modality. Hence, we introduce two baselines: 1) caption the image with Instruct BLIP \cite{instructBLIP} and then pass it along with the caption to MusicLM \cite{musiclm}. We call this baseline \textbf{MusicLM + InstructBLIP}.  2) we adapt a recent text-to-audio diffusion model TANGO \cite{tango} into our setting, as explained next. We call its modified version TANGO++. Further, we compare ourselves with $7$ other text-to-music approaches. We elaborate on them below:

% To the best of our knowledge we propose the first multi-modal music generation pipeline. To this end, we propose a baseline:
% \noindent{\textbf{Aligning the Image-Text modalities through Contrastive Loss.}}
% \noindent{\textbf{MusicLM + InstructBLIP :}}

\noindent{\textbf{TANGO++:}}
TANGO \cite{tango} is a powerful text-to-audio model based on LDMs. They condition the diffusion model on text embeddings from FLAN-T5 \cite{flant5} text encoder $\bm{z}_{text}$. To facilitate joint conditioning from text and image $\bm{I}$, we embed $\bm{I}$ to the latent space as $\bm{z}_{image}$ using a ViT \cite{vit} based CLIP encoder, and align them together through an Image-Text Contrastive loss. Once they are aligned, both the embeddings are fused and the LDM is jointly conditioned.

\noindent{\textbf{Text-to-Music Baselines:}}
To bring out the utility of conditioning on both visual and textual modality, we compare \modelname with seven other text-to-music methods too: Riffusion \cite{riffusion}, Mubert \cite{mubert}, MusicLM \cite{musiclm}, Moûsai \cite{mousai}, Noise2Music \cite{noise2music}, MeLoDy \cite{melody} and MusicGen \cite{musicgen}. We provide more details of each of these in the Appendix.

\subsection{Results} \label{sec:main_results}
We present exhaustive objective and subjective comparison of \modelname with the baseline approaches in \cref{tab:main_table}.
When compared with text-to-music approaches in the first section of the table, our results show the significant utility of adding extra visual conditioning on the generations. The fine-grained contextual information from visual modality is able to supplement the information from the corresponding text, thereby enhancing the quality of music generation.
Further, \modelname is able to consistently outperform MusicLM + InstructBLIP and TANGO++ (which has similar conditioning as ours). This highlights the efficacy of our visual synapse, which infuses the right amount of visual conditioning to enable the model to synthesize perceptually congruent music tracks. Captions from InstructBLIP are superfluous and vague when compared to expert-annotated, high-quality MusicCaps captions on which MusicLM is trained. This distributional shift leads to a performance drop as shown in the table. TANGO++ uses contrastive loss to align CLIP image features and FLAN-T5 text features, further, we use simple addition for joint conditioning -- these design choices can be sub-optimal. 

Our subjective human evaluation in \cref{tab:main_table} also suggests that conditioning the music generation on both visual and textual modality improves its perceptual quality.

\section{Discussions and Analysis}

% \noindent{\textbf{Sensitivity analysis on DDIM inversion}} \\

% \begin{table}
% \centering
% \resizebox{\columnwidth}{!}
% {\begin{tabular}{c|c|c|c|c|c}
% \toprule
% \multirow{2}{*}{ \bf \# Denoising Steps } & \multicolumn{3}{|c|}{ \bf Objective metrics } & \multicolumn{2}{c}{ \bf Subjective metrics } \\
% \cmidrule { 2 -6 }
% & \textbf{FD} & \textbf{KL} & \textbf{FAD} & \textbf{OVL} & \textbf{REL} \\
% \midrule
% - & 25.97 & 1.01 & 4.00 & & \\
% \CC{}X & \CC{}\textbf{22.65} & \CC{}\textbf{0.89} & \CC{}\textbf{1.12} & \CC{} & \CC{} \\
% \bottomrule
% \end{tabular}}
% \caption{Ablation 1}
% \label{tab:ablation_1}
% \end{table}

% \noindent{\textbf{Effect of image conditioning}} Inference on extent of image information that is getting injected into text modality \\

% \noindent{\textbf{Impact of $\alpha$ Conditioning.}} 

\customsubsection{Analyzing the Design Choice of $\alpha$ parameters:} 
$\alpha$ parameters introduced in \cref{eqn:synapse} controls how the self-attention features from the blocks within the text-to-image diffusion model interact with the cross-attention features from the text-to-music diffusion model. \modelname has one alpha parameter per block within the decoders of both diffusion models. We vary this design choice in \cref{tab:role_of_alpha}. Attaching the synapse in the decoder offers better performance. This is because the decoder controls the major transformations that contribute to generating the image. Further, learning different $\alpha$ per block helps to learn block-specific mixing co-efficient, which slightly improves the performance. 

We also perform a sensitivity analysis on the learning rate (LR) used while learning $\alpha$ parameters in \cref{tab:alpha_lr}. Based on this analysis, we use a LR of $1e-5$ in our experiments.

\begin{table}
\centering
\resizebox{\columnwidth}{!}
{\begin{tabular}{c c|c c c|c c c}
\toprule
\multirow{2}{*}{ \bf Encoder } & 
\multirow{2}{*}{ \bf Decoder } & \multicolumn{3}{|c|}{ \bf Extended MusicCaps} & \multicolumn{3}{c}{ \bf \ourdataset } \\
\cmidrule { 3 - 8 }
& & \textbf{FAD} $\downarrow$ & \textbf{KL} $\downarrow$ & \textbf{FD} $\downarrow$ & \textbf{FAD} $\downarrow$ & \textbf{KL} $\downarrow$ & \textbf{FD} $\downarrow$ \\
\midrule
\multicolumn{8}{c}{\textit{Same $\alpha$ for all blocks.}}\\
\midrule
\textcolor{ForestGreen}{\ding{51}} & \textcolor{OrangeRed}{\ding{55}} & 3.22 & 1.23 & 24.01 & 2.01 & 1.01 & 21.96 \\
\textcolor{ForestGreen}{\ding{51}} & \textcolor{ForestGreen}{\ding{51}} &  2.71 & 1.13 & 23.31 & 1.27 & 0.87 & 21.04 \\
\textcolor{OrangeRed}{\ding{55}} & \textcolor{ForestGreen}{\ding{51}} &  2.79 & 1.14 & 23.44 & 1.29 & 0.87 & 21.13 \\
\midrule
\multicolumn{8}{c}{\textit{Different $\alpha$ for each block.}}\\
\midrule
\textcolor{ForestGreen}{\ding{51}} & \textcolor{OrangeRed}{\ding{55}} & 2.03 & 1.10 & 23.36 & 1.92 & 0.93 & 21.28 \\
\textcolor{ForestGreen}{\ding{51}} & \textcolor{ForestGreen}{\ding{51}} &  1.13 & 0.94 & 22.81 & 1.07 & 0.76 & 20.53 \\
\CC{}\textcolor{OrangeRed}{\ding{55}} & \CC{}\textcolor{ForestGreen}{\ding{51}} & \CC{}\textbf{1.12} & \CC{}\textbf{0.89} & \CC{}\textbf{22.65} & \CC{}\textbf{1.05} & \CC{}\textbf{0.72} & \CC{}\textbf{20.49} \\
\bottomrule
\end{tabular}}
\caption{
We systematically analyze our design choice of learnable $\alpha$ parameters. We vary the position of the synapse: encoder or decoder and also study whether we need the same or different $\alpha$ parameters for each block within them. 
% Impact of $\alpha$ learning across different layers. We study two variants: (i) Same $\alpha$ learned for all layers and (ii) Different $\alpha$ for each layer over all possible combinations of Encoder and Decoder. 
}
\label{tab:role_of_alpha}
\end{table}

\begin{table}
\centering
\resizebox{0.95\columnwidth}{!}
{\begin{tabular}{c|c c c|c c c}
\toprule
\multirow{2}{*}{ \bf Learning Rate } & \multicolumn{3}{|c|}{ \bf Extended MusicCaps} & \multicolumn{3}{c}{ \bf \ourdataset } \\
\cmidrule { 2 - 7 }
& \textbf{FAD} $\downarrow$ & \textbf{KL} $\downarrow$ & \textbf{FD} $\downarrow$ & \textbf{FAD} $\downarrow$ & \textbf{KL} $\downarrow$ & \textbf{FD} $\downarrow$ \\
\midrule
0.5e-6 & 3.12 & 1.21 & 23.26 & 2.01 & 1.14 & 21.95 \\
0.5e-4 & 1.38 & 0.95 & 22.81 & 1.39 & 0.88 & 20.86 \\
1e-6 & 2.56 & 1.17 & 23.11 & 1.86 & 1.10 & 21.71 \\
\CC{}\textbf{1e-5} & \CC{}\textbf{1.12} & \CC{}\textbf{0.89} & \CC{}\textbf{22.65} & \CC{}\textbf{1.05} & \CC{}\textbf{0.72} & \CC{}\textbf{20.49} \\
\bottomrule
\end{tabular}}
\caption{
Sensitivity analysis on the learning rate for $\alpha$ parameters.
% Ablation on different learning rates of $\alpha$ on the two datasets for music generation task.
}
\label{tab:alpha_lr}
\end{table}

% \noindent{\textbf{Performance on Different Input Conditioning.}}
\customsubsection{Efficacy of Conditioning on both Modalities:}
In order to study the contribution of each modality on \modelname, we train three different variations of the model by selectively turning off visual conditioning and textual conditioning. We report the results in \cref{tab:input_conditioning}. We see significant improvement when conditioning on both modalities. 
This highlights how complementary semantic information from each modality can compose better music.

% We evaluate \modelname~ under different input conditioning as reported in Table \ref{tab:input_conditioning}. Note that our model when equipped with both image and text prompts performs better than its image-only, and text-only counterparts. Multi-conditioned prompting leverages the synergy between the two complementary modalities to capture the fine-grained inter-modal semantics which results in perceptually realistic and high-fidelity soundtracks. We regulate Equation \ref{eq:1} to control the impact of individual modalities.      

% \noindent{\textbf{Performance on Different Input Conditioning.}}
% We evaluate \modelname~ under different input conditioning as reported in Table \ref{tab:input_conditioning}. Note that our model when equipped with both image and text prompts performs better than its image-only, and text-only counterparts. Multi-conditioned prompting leverages the synergy between the two complementary modalities to capture the fine-grained inter-modal semantics which results in perceptually realistic and high-fidelity soundtracks. We regulate Equation \ref{eq:1} to control the impact of individual modalities.      

\begin{table}
\centering
\resizebox{\columnwidth}{!}
{\begin{tabular}{c c|c c c|c c c}
\toprule
\multirow{2}{*}{ \bf Text } & 
\multirow{2}{*}{ \bf Image } & \multicolumn{3}{|c|}{ \bf Extended MusicCaps} & \multicolumn{3}{c}{ \bf \ourdataset } \\
\cmidrule { 3 - 8 }
& & \textbf{FAD} $\downarrow$ & \textbf{KL} $\downarrow$ & \textbf{\imagemusicmetric } $\uparrow$ & \textbf{FAD} $\downarrow$ & \textbf{KL} $\downarrow$ & \textbf{\imagemusicmetric} $\uparrow$ \\
\midrule
\textcolor{ForestGreen}{\ding{51}} & \textcolor{OrangeRed}{\ding{55}} & 3.07 & 1.21 & - & 3.11 & 1.19 & - \\
\textcolor{OrangeRed}{\ding{55}} & \textcolor{ForestGreen}{\ding{51}} & 5.62 & 1.54 & - & 4.16 & 1.37 & - \\
\CC{}\textcolor{ForestGreen}{\ding{51}} & \CC{}\textcolor{ForestGreen}{\ding{51}} & \CC{}\textbf{1.12} & \CC{}\textbf{0.89} & \CC{}\textbf{0.76} & \CC{}\textbf{1.05} & \CC{}\textbf{0.72} & \CC{}\textbf{0.83} \\
\bottomrule
\end{tabular}}
\caption{
Conditioning independently on each of the modalities leads to inferior music generation performance in this experiment. 
% Result comparison of \modelname~ when prompted with different modalities e.g., only-image, only-text, image-text both.
}
\label{tab:input_conditioning}
\end{table}

% \noindent{\textbf{Dual Diffusion}} Comparison against MICAI paper (\textcolor{red}{Sayan}) [link added in doc]

% \noindent{\textbf{Analysis on Varying Inference Steps and CFG}}
\customsubsection{Sensitivity Analysis:}
\cref{tab:steps_and_guidance} reports the sensitivity analysis of changing the number of denoising steps $T$ and the strength of classifier-free guidance during inference. Similar to the findings from \citet{tango}, increasing $T$ helps to generate more pleasing music. This can be attributed to enhanced refinement from more denoising. CFG strength of $7$ gives the best result, and we use it in our experiments. 

% The number of inference steps and the classifier-free guidance scale play a critical role during sampling from latent diffusion models. We report the effect of these two aspects on music piece generation in Table \ref{tab:steps_and_guidance}. For our experiments, a guidance scale of 7 achieves the optimal results for \modelname. As shown in the table, the guidance scale is fixed to 7 across a varying number of steps ranging from $20$ to $400$. We see that the objective metrics become gradually better with increasing steps. We also report the effect of varying guidance scales with fixed steps $400$. Having very low CFG produces suboptimal results across all the objective measures. The best FAD metric is attained at a guidance scale of 7 with subsequent drops in values with larger guidance.

\begin{table}
\centering
\resizebox{\columnwidth}{!}
{\begin{tabular}{c|cccc|c|cccc}
\toprule
\multicolumn{5}{c|}{\bf Varying Steps} & \multicolumn{5}{c}{ \bf Varying Guidance } \\
\midrule
\textbf{Guidance} & 
\textbf{Steps} &
\textbf{FAD}$\downarrow$ & \textbf{KL}$\downarrow$ & \textbf{FD}$\downarrow$ & \textbf{Steps} & \textbf{Guidance} & \textbf{FAD}$\downarrow$ & \textbf{KL}$\downarrow$ & \textbf{FD}$\downarrow$  \\
\midrule
% \multirow{5}{*}{7}
% & 20 & 26.68 & 1.26 & 1.41 & \multirow{5}{*}{400}
% & 2 & 24.64 & 1.13 & 1.47\\
\multirow{5}{*}{7}
& 50 & 2.59 & 1.97 & 27.45 & \multirow{5}{*}{400}
& 2 & 1.47 & 1.13 & 24.64 \\
% & 100 & 25.89 & 1.24 & 1.39 & & 5 & 23.31 & 0.98 & 1.29 \\
& 200 & 1.35 & 1.12 & 25.22 & & 7 & \textbf{1.12} & \textbf{0.89} & \textbf{22.65} \\
% & 300 & 23.97 & 1.03 & 1.27 & & 15 & 23.44 & 1.04 & 1.35 \\
& 400 & \underline{1.12} & \underline{0.89} & \underline{22.65} & & 20 & 1.51 & 1.18 & 24.92 \\
& 600 & 1.09 & 0.88 & 22.57 & & 30 & 1.63 & 1.29 & 25.38 \\
& 800 & 1.07 & 0.77 & 22.48 & & 50 & 1.58 & 1.34 & 24.87 \\
\bottomrule
\end{tabular}}
\caption{
Sensitivity analysis on the number of denoising steps $T$, and the strength of classifier-free guidance.
% \modelname~ performance with varying inference steps and classifier-free guidance
}
\label{tab:steps_and_guidance}
\end{table}

\begin{table}
\centering
\resizebox{\columnwidth}{!}
{\begin{tabular}{c|c|c c c | c c}
\toprule
\multirow{2}{3.0 cm}{ \centering \bf Text prompt length (in words) } & \multirow{2}{*}{ \bf Image } & \multicolumn{3}{|c|}{ \bf Objective metrics} & \multicolumn{2}{c}{ \bf Subjective metrics } \\
\cmidrule { 3 - 7 }
& & \textbf{FAD} $\downarrow$ & \textbf{KL} $\downarrow$ & \textbf{FD} $\downarrow$ & \textbf{OVL} $\uparrow$ & \textbf{REL} $\uparrow$ \\
\midrule
$\ge$ 8 $\le$ 13 & \textcolor{OrangeRed}{\ding{55}} & 5.28 & 1.35 & 25.81 & 82.86 & 82.54 \\
$\ge$ 14 $\le$ 19 & \textcolor{OrangeRed}{\ding{55}} & 3.13 & 1.20 & 23.11 & 85.25 & 85.16 \\
$\ge$ 20 & \textcolor{OrangeRed}{\ding{55}} & 3.02 & 1.19 & 22.65 & 86.04 & 85.96 \\
\CC{}\textbf{$\le$ 7} & \CC{}\textcolor{ForestGreen}{\ding{51}} & \CC{}\textbf{1.86} & \CC{}\textbf{0.87} & \CC{}\textbf{21.36} & \CC{}\textbf{87.13} & \CC{}\textbf{86.21} \\
\bottomrule
\end{tabular}}
\caption{
% Analyzing whether more verbose text prompts can sidestep image conditioning. 
Performance of \modelname~ with varying verbosity of text prompts collected from \ourdataset.
}
\label{tab:varying text prompt}
\end{table}

% \noindent{\textbf{Benefits of Image Fusion.}}
% importance of image. highlight with small text prompt how our method works well
\customsubsection{Verbose Text versus Image Conditioning:}
The visual synapse infuses fine-grained semantics from the image into text-to-music diffusion models. Another alternative to infuse such semantics would be to use verbose text descriptions. To study this, we remove the visual synapse from \modelname and train a music generation model conditioned only on text. Then, we vary the length of text prompts and report results in \cref{tab:varying text prompt}. Interestingly, we find that using image conditioning with a small text prompt outperforms using lengthier prompts. This underscores the utility of visual conditioning and the ability of visual synapses to modulate the LDM effectively.

% To study the importance of the information injected by the image branch, we experiment with varying verbosity of the text prompt. Table \ref{tab:varying text prompt} reports the performance of \modelname~ when presented with textual descriptions with different sparsity. It is to be noted that the model variant with a very short description (up to $5$ words) with an image supplement outperforms when the textual description is considerably longer ($\ge 20$ words) underlining the significance of the visual prompt. The shortest text prompt (up to 7 words) consisted of examples like (a) \textit{`A person playing guitar'} (b) \textit{`Jazz music performance'} (c) \textit{`Haunting sound'} or simply (d) \textit{`Concert'}. Note, all these cases leave room for ambiguities like in (a) is the guitar acoustic or electric? (b) Does the Jazz composition contain a solitary Piano vs. a combination of Saxophone, Clarinet, Trumpet, and Cello (c) Is the backdrop set in a dark alley vs inside a spooky forest? (d) Is it a rock concert or a classical concert? To this end, our experiments underline the importance of the addition of the visual modality.

\customsubsection{Effect of visual-cues:} We analyse the effect of using a different image and the same text prompt (please refer to the project page). When we change from the walkway image to the blue forest, the music becomes more calm and distant. We change the image to an abandoned amusement park, carnival orchestra, foggy seaside concert and forest at night. We observe a prevalence of eerie ambient sound echoing through the deserted park, occasional circus-inspired motifs, distant sounds of waves and foghorn-like effects and atmospheric strings imitating rustling leaves respectively.

\customsubsection{Comparison with Text-to-Audio Methods:} We include a comparison of \modelname with text-to-audio generation approaches in \cref{tab:comparison_against_text_to_audio}. We finetune their pre-trained checkpoints on our benchmark datasets for this experiment. The complementary information from both modalities allows our approach to outperform these methods too.
% We observe that \modelname outperforms prior methods by a considerable margin. This can be attributed to our mode fine-grained semantic understanding in terms of both modalities. 

% \noindent{\textbf{Comparison against Text-to-Audio methods.}} We compare our method against benchmark text-to-audio generation methods in \cref{tab:comparison_against_text_to_audio}. We leverage their pre-trained model and finetune on our datasets. We observe that \modelname outperforms prior methods by a considerable margin. This can be attributed to our model's fine-grained semantic understanding in terms of both modalities.    

\begin{table}
\centering
\resizebox{\columnwidth}{!}
{\begin{tabular}{c|c c c|c c c}
\toprule
\multirow{2}{*}{ \bf Method } & \multicolumn{3}{|c|}{ \bf MusicCaps} & \multicolumn{3}{c}{ \bf \ourdataset } \\
\cmidrule { 2 - 7 }
& \textbf{FAD} $\downarrow$ & \textbf{KL} $\downarrow$ & \textbf{FD} $\downarrow$ & \textbf{FAD} $\downarrow$ & \textbf{KL} $\downarrow$ & \textbf{FD} $\downarrow$ \\
\midrule
% AudioGen \cite{audiogen} & - & - & - & - & - & - \\
AudioLDM \cite{audioldm} & 2.29 & 1.29 & 24.07 & 1.86 & 1.42 & 22.49 \\
TANGO \cite{tango} & 1.96 & 1.17 & 23.31 & 1.93 & 1.18 & 21.92 \\ \midrule
\CC{}\textbf{\modelname} & \CC{}\textbf{1.12} & \CC{}\textbf{0.89} & \CC{}\textbf{22.65} & \CC{}\textbf{1.05} & \CC{}\textbf{0.72} & \CC{}\textbf{20.49} \\
\bottomrule
\end{tabular}}
\caption{
While comparing \modelname with state-of-the-art text-to-audio approaches, we see significant improvement in quality.
% comparison against text to audio methods
}
\label{tab:comparison_against_text_to_audio}
\end{table}

\customsubsection{Effectiveness of \imagemusicmetric:} We conduct a user study with $64$ participants to check whether the proposed \imagemusicmetric metric is indeed capturing the relatedness between generated music and the conditioning image. We randomly choose $300$ samples from Extended MusicCaps and \ourdataset each. We compute the \imagemusicmetric score, and also ask users for their image-music similarity on a scale of [0,1], for these samples. The average score from \imagemusicmetric metric and the users for Extended MusicCaps and \ourdataset are ($0.76$, $0.71$) and ($0.83$, $0.85$) respectively. The high correlation underscores the validity and usefulness of \imagemusicmetric metric.

\customsubsection{Using \imagemusicmetric to Measure Purity of the Datasets:}
We compute the IMSM scores for all image-music pairs present in both Extended MusicCaps and \ourdataset datasets and obtain a score of $0.91$ and $0.93$ respectively. The purpose of this is to quantitatively establish that the curated samples are perceptually in sync and are meaningful. The high values of the IMSM scores demonstrate that the curated image samples are highly perceptually similar and have ample association with the musical compositions.

% To examine the correlation of our proposed metric \imagemusicmetric~ with human perception, we perform a user study analysis. From the extended MusicCaps and \ourdataset~ test set we randomly choose $300$ samples and compute the mean \imagemusicmetric~ score (on a scale of 0-1). We simultaneously ask the users taking the survey to rate the image-music similarity on a scale of [0,1]. Fig \ref{fig:cimp-relevance} supports our hypothesis as we find the \imagemusicmetric~ to have high association with human subject scores underlining the validity and usefulness of such a metric in this task. 

% \input{sec/6_conclusions}
\vspace{10pt}
\section{Conclusion and Future Works}
\label{Conclusions}
We explore the utility of infusing image semantics into a text-to-music diffusion model, enabling us to generate music, consistent with both visual and textual semantics in this work. To the best of our knowledge, ours is the first effort towards such a multi-conditioned music generation. We develop \modelname with a novel ``visual synapse" to effectively infuse the image semantics into music generation, introduce a new dataset \ourdataset, and propose a new evaluation metric.
We conduct exhaustive experimental analysis on  \ourdataset and a modified version of MusicCaps \cite{musiclm} and compare \modelname against $7$ text-to-music methods, and a modified baseline. The results suggest: $1$) the extra information from the image conditioning significantly boosts music generation quality $2$) our ``visual synapse" is effective in modulating and infusing the required semantic information into the generative process.

\modelname can be an essential tool for a creative professional or a social-media content creator who needs to generate music that can go well with their multi-modal post (consider a user posting about their recent picnic -- their photos can be the image conditioning while a short description of the trip can be the textual input to \modelname). Creating music with semantic lyrics that can go well with a video can be some interesting open-ended follow-up works.

{
    \small
    \bibliographystyle{ieeenat_fullname}
    \bibliography{main}
}

\newpage

% \title{\modelname: Synthesizing \underline{M}usic from Imag\underline{e} and \underline{L}anguage Cues using \\ Dif\underline{fusion} Models\\ \textcolor{blue}{Appendix}}

% \title{\modelname: Synthesizing \underline{M}usic with Imag\underline{e}-\underline{L}anguage Conditioned\\ Dif\underline{fusion} Model}
% \title{Synthesizing Music from Image and Language Cues using Diffusion Models}

%%%%%%%%% AUTHORS - PLEASE UPDATE
% \and

% \newpage

% \begin{document}
% \maketitle
% \thispagestyle{empty}
% \appendix

\newpage
\appendix

% \section{Appendix / supplemental material}
\begin{center}
\begin{LARGE}
\textbf{\underline{\textcolor{blue}{Appendix}}}
% \vspace{4mm}
\end{LARGE}
\end{center}

\noindent In this appendix we provide additional information on the following: \\
\ref{sec:more on tango++} More Details on TANGO++ \\
\ref{sec:problem motivation} Problem Motivation Revisited \\
\ref{sec: other baselines} Other Baseline Approaches\\
\ref{sec: implementation details} Implementation Details \\
\ref{sec: more experiments} More Experimental Analysis \\
\ref{sec: more dataset details} Dataset Details \\
\ref{sec: more user study} User Study Details \\
\ref{sec: conditional image gen} Inspiration from Conditional Image Generation \\
\ref{sec: related audio concepts} Related Audio Concepts \\

% \subsection{TANGO++:}
\section{More Details on TANGO++}
\label{sec:more on tango++}
Our modified baseline model TANGO++ comprises an early-fusion approach, where we align the visual and the textual
modalities through an Image-Text Contrastive (ITC) loss. As the generated music is conditioned on both modalities, bringing them to a common latent space is imperative to the success of the system. The text input is passed through the FLAN-T5 text encoder which we keep as frozen. For image encoding
we use ViT \cite{vit}. We project the visual and the textual inputs to a common embedding space and align them using ITC loss. The diffusion model is conditioned on this hybrid embedding to produce audio signals. It is then converted into spectrograms using the decoder and then passed through a HiFi GAN vocoder to produce the music signal. The expression for ITC loss ($\mathcal{L}_{\text{ITC}}$) is as follows:

\begin{align}\label{eq:clip}
\mathcal{L}_{\text{ITC}} = -\frac{1}{2\mathcal{N}}\sum_{j = 1}^{\mathcal{N}}\log\underbrace{\left[\frac{\exp\left({\langle}z^{I}_{j}, z^{T}_{j}{\rangle}/\tau\right)}{\sum_{l = 1}^{\mathcal{N}}{\exp\left({\langle}z^{I}_{j}, z^{T}_{l}{\rangle}/\tau\right)}}\right]}_{\text{Contrasting images with the texts}} \nonumber \\ -\frac{1}{2\mathcal{N}}\sum_{l = 1}^{\mathcal{N}}\log\underbrace{\left[\frac{\exp\left({\langle}z^{I}_{l}, z^{T}_{l}{\rangle}/\tau\right)}{\sum_{j = 1}^{\mathcal{N}}{\exp\left({\langle}z^{I}_{j}, z^{T}_{l}{\rangle}/\tau\right)}}\right]}_{\text{Contrasting texts with the images}}
\end{align}

where $\langle \cdot, \cdot\rangle$ denotes inner product, and $\tau$ is the temperature parameter. $z^I$ and $z^T$ refer to the image and text latent representations respectively.

% \vspace{-0.26in}

\section{Problem Motivation Revisited}
\label{sec:problem motivation}
\begin{figure}[H]
    \centering
\includegraphics[width=0.29\textwidth]{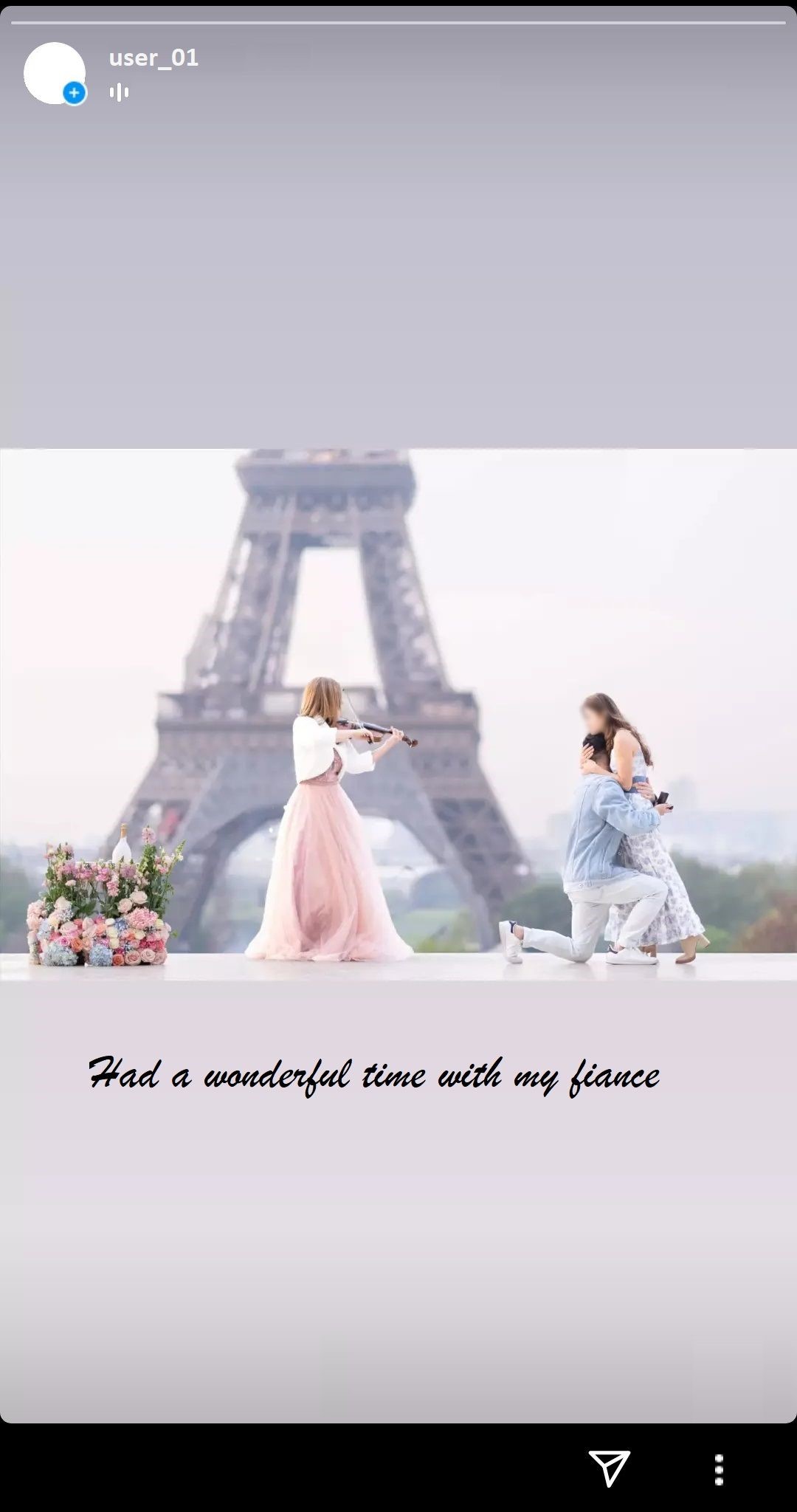}
    \caption{A mock-up of a  social media post that contains an image and associated textual content. Our approach \modelname, can consume such image-textual pairs as input and synthesize music that can go well with them.}
    \label{fig:problem-motivation}
\end{figure}

Social media platforms have become ubiquitous and provide a channel for everyone to express their creativity and share their happenings with the world. It is very common for users to upload an image, and write an associated text with it (\cref{fig:problem-motivation}). Adding music to these social media posts enhances its visibility and appeal. Instead of retrieving music from an existing database, our approach \modelname, will be able to generate music tracks that are custom-made, conditioned on the uploaded image and its description. We note that ours is the first approach that operates in this pragmatic setting, to generate music conditioned on both visual and textual modality.

% \vspace{-1mm}

\section{Other Baseline Approaches}
\label{sec: other baselines}
In addition to our proposed baseline approach, we compare \modelname against the following methods. Note that these are text-to-music generation methods unlike our approach and don't support multi-conditioning in input prompts. Hence a direct comparison might not be entirely fair. In most cases these methods don't support introducing an additional modality conditioning as a result we compare our approach against these baselines directly to study the benefits of \modelname.

Riffusion \cite{riffusion} base their algorithm on fine-tuning a Stable Diffusion model \cite{rombach2022high} on mel spectrograms of music pieces from a paired music-text dataset. This is one of the first text-to-music generation methods. Mubert \cite{mubert} is an API-based service that employs a Transformer backbone. The encoded prompt is used to match the music tags and the one with the highest similarity is used to query the audio generation API. It operates over a relatively smaller set as it produces a combination of audio from a predefined collection. MusicLM \cite{musiclm} generates high-fidelity music from text descriptions by casting the process of conditional music generation as a hierarchical sequence-to-sequence modeling task. They leverage the audio-embedding network of MuLan \cite{mulan} to extract the representation of the target audio sequence. Moûsai \cite{mousai} is a cascading two-stage latent diffusion model that is equipped to produce long-duration high-quality stereo music. It achieves this by employing a specially designed U-Net facilitating a high compression rate. Noise2Music \cite{noise2music} introduced a series of diffusion models, a generator, and a cascader model. The former generates an intermediate representation
conditioned on text, while the later can produce audio conditioned on the intermediate representation of the text. MeLoDy \cite{melody} pursues an LM-guided diffusion model by reducing the forward pass bottleneck and applies a novel dual-path diffusion mode. MusicGen \cite{musicgen} comprises a single-stage transformer LM together with efficient token interleaving patterns. This eliminates the need for hierarchical upsampling.

\section{Implementation Details} \label{sec: implementation details}
Our text-to-music LDM contains $3$ encoder blocks and $3$ decoder blocks, similar to \citet{tango}. Empirically we find that finetuning from its pre-trained checkpoint helps convergence. FLAN-T5 \cite{flant5} is used as the text encoder. \modelname is trained for $30$ epochs using AdamW optimizer \cite{adamw}. We attach our visual synapse only on the decoder layers of the LDM. Similar to earlier works \cite{audioldm,tango}, we find that using classifier-free guidance improves the result. Our training takes $42$ hours on $4$ NVIDIA A$100$ GPUs.

\section{More Experimental Analysis}
\label{sec: more experiments}

\subsection{Choice of Text-to-Image Diffusion Model}

\begin{table}[H]
\centering
\resizebox{\columnwidth}{!}
{\begin{tabular}{c|c c c|c c c}
\toprule
\multirow{2}{*}{ \bf Model } & \multicolumn{3}{|c|}{ \bf MusicCaps} & \multicolumn{3}{c}{ \bf \ourdataset } \\
\cmidrule { 2 - 7 }
& \textbf{FD} $\downarrow$ & \textbf{KL} $\downarrow$ & \textbf{FAD} $\downarrow$ & \textbf{FD} $\downarrow$ & \textbf{KL} $\downarrow$ & \textbf{FAD} $\downarrow$ \\
\midrule
Stable Diffusion V1.2 & 1.84 & 1.52 & 22.88 & 1.49 & 1.14 & 21.44 \\
Stable Diffusion V1.3 & 1.62 & 1.29 & 22.72 & 1.34 & 1.03 & 21.02 \\
Stable Diffusion V1.4 & 1.31 & 1.13 & 22.67 & 1.20 & 0.91 & 20.53 \\
\CC{}\textbf{Stable DiffusionV1.5} & \CC{}\textbf{1.12} & \CC{}\textbf{0.89} & \CC{}\textbf{22.65} & \CC{}\textbf{1.05} & \CC{}\textbf{0.72} & \CC{}\textbf{20.49} \\
\bottomrule
\end{tabular}}
\caption{\modelname with different versions of Stable Diffusion.}
\label{tab:sd_version}
\end{table}
We study the effect of employing different variants of the text-to-image Stable Diffusion model (V1.2 through V1.5) in \cref{tab:sd_version}. We note that the best results are obtained with the latest variant. This brings to light that our proposed visual synapse is able to cascade the usage of better text-to-image models into improving the quality of music generation. The Stable Diffusion V1.4 and V1.5 checkpoints were initialized with the weights of the Stable Diffusion V1.2 checkpoint and subsequently fine-tuned on 225k steps at resolution $512 \times 512$ on the LAION dataset and 10\% dropping of the text-conditioning to improve classifier-free guidance sampling.

\subsection{Performance with Different Text Encoders}

\begin{table}[H]
\centering
\resizebox{\columnwidth}{!}
{\begin{tabular}{c|c c c|c c c}
\toprule
\multirow{2}{*}{ \bf Model } & \multicolumn{3}{|c|}{ \bf MusicCaps} & \multicolumn{3}{c}{ \bf \ourdataset } \\
\cmidrule { 2 - 7 }
& \textbf{FAD} $\downarrow$ & \textbf{KL} $\downarrow$ & \textbf{FD} $\downarrow$ & \textbf{FAD} $\downarrow$ & \textbf{KL} $\downarrow$ & \textbf{FD} $\downarrow$ \\
\midrule
BERT \cite{bert} & 2.82 & 2.23 & 24.73 & 2.91 & 1.94 & 22.13 \\
RoBERTa \cite{roberta} & 2.35 & 2.02 & 24.09 & 2.17 & 1.87 & 21.95 \\
T5-Small \cite{t5}  & 1.98 & 1.79 & 23.68 & 1.89 & 1.66 & 21.23 \\
T0 \cite{t0} & 1.42 & 1.25 & 22.96 & 1.32 & 1.19 & 20.76 \\
% PalLM & - & - & - & - & - & - \\
CLIPText & 1.24 & 0.94 & 22.78 & 1.16 & 0.91 & 20.58 \\
\CC{}\textbf{FLAN-T5 \cite{flant5}} & \CC{}\textbf{1.12} & \CC{}\textbf{0.89} & \CC{}\textbf{22.65} & \CC{}\textbf{1.05} & \CC{}\textbf{0.72} & \CC{}\textbf{20.49} \\
\bottomrule
\end{tabular}}
\caption{Performance of \modelname with different text encoders}
\label{tab:text_encoders}
\end{table}
In \cref{tab:text_encoders} we compare the performance of \modelname under different text encoders. We note that the best results are achieved when an instruction-tuned text encoder is employed (FLAN-T5 \cite{flant5}) over other non-instruction-based models, which correlates with the findings in \citet{tango}. This is very closely followed by the ClipText \cite{clip} encoder.

% \noindent{\textbf{Implementation details}}
% We leverage the pre-trained model from AudioLDM \citet{audioldm} and finetune it on \ourdataset and Extended MusicCaps. We use a frozen Flan-T5 as the text encoder due to its superior performance by virtue of being instruction-tuned. We train the model for 30 epochs and report results using the checkpoint with the best validation loss. We use AdamW optimiser \cite{adamw}. We train the model on 4 A100 GPUS with validation at the end of each epoch. Training our model takes around 42 hours for 30 epochs with gradient accumulation. 

% We freeze the FLAN-T5-LARGE text encoder and only train the parameters of the latent diffusion model. The diffusion model is based on the Stable Diffusion U-Net architecture. We use 8
% channels and a cross-attention dimension of 1024 in the U-Net model. We use the AdamW optimizer \cite{} with a learning rate of 1e-5 and a linear learning rate scheduler
% for training. We train the model for 40 epochs on the two datasets and report results for the checkpoint with the best validation loss, which we obtained at epoch 39. We use four A100 GPUs for training our model, where it takes a total of 36 hours to train 35 epochs, with validation at the end of every epoch. We use a per GPU batch size of 3 (2 original + 1 augmented instance) with 4 gradient accumulation steps. The effective batch size for training is 3 (instance) ∗4 (accumulation) ∗4 (GPU) = 48.

\subsection{Variation Across Genres}

\begin{table}[H]
\centering
\resizebox{\columnwidth}{!}
{\begin{tabular}{c|c c c c | c c}
\toprule
\multirow{2}{*}{ \bf Genre name } & \multicolumn{4}{|c|}{ \bf Objective metrics} & \multicolumn{2}{c}{ \bf Subjective metrics } \\
\cmidrule { 2 - 7 }
& \textbf{FD} $\downarrow$ & \textbf{KL} $\downarrow$ & \textbf{FAD$_{\text {VGG}}$} $\downarrow$ & \textbf{\imagemusicmetric} $\uparrow$ & \textbf{OVL} $\uparrow$ & \textbf{REL} $\uparrow$ \\
\midrule
Pop & 22.47 & 0.78 & 1.21 & 0.95 & 86.31 & 90.10 \\
Rock & 21.11 & 0.95 & 0.85 & 0.81 & 88.41 & 84.92 \\
Hip-Hop/Rap & 19.73 & 0.65 & 1.24 & 0.69 & 83.05 & 88.78 \\
Electronic Dance Music & 20.03 & 1.06 & 0.93 & 0.72 & 85.39 & 86.18 \\
Country & 19.56 & 0.89 & 0.88 & 0.98 & 89.94 & 87.22 \\
\bottomrule
\end{tabular}}
\caption{A study on the diversity analysis of \modelname. We evaluate the performance of our model on generating musical tracks of five different genres on \ourdataset. }
\label{tab:genre_wise_performance}
\end{table}

% discuss subjective results by subject matter experts 
Tab \ref{tab:genre_wise_performance} reports the performance of \modelname~ across the $5$ most popular genres (chosen through a study undertaken by \cite{genre_popularity}) on the genre-wise test set collected from \ourdataset. We find a steady performance of our approach across different genres substantiating the ability of the model to capture the musical nuances like the composition of the instruments, track progression, sequence of instruments introduced, rhythm, tonality, tempo, and beats. Due to the highly subjective nature of the problem, we also perform a human evaluation by subject matter experts. To this end, we employ $7$ individuals formally trained in music to independently listen and report OVL and REL scores considering the aforementioned aspects to assess the quality of genre-wise samples. We report the mean OVL and REL values from all the evaluators on a subset of the corresponding genre-wise test splits. We find that the overall performance of our method is highly encouraging as reported in Tab \ref{tab:genre_wise_performance}.  

\subsection{Ablating choice of layers}
When we fuse subset of Decoder Blocks, we see drop in performance in Tab. \ref{tab:decoder_block_ablation_rebuttal}, as coupling becomes weak. We also ablate encoder and decoder layer separately (refer to Tab. 2 of main paper). Learned $\alpha$ values for each blocks ($0.37$, $0.59$ and $0.63$ respectively) improves over $\alpha\text{=}0.5$ on all metrics, thus avoiding an extra hyper-parameter to tune. With a few layers to account for dimension mismatch, visual synapse can scale to different architectures and avoid layer-to-layer correspondence. We will explore this in a future work.

\begin{table}[H]
\centering
\resizebox{\columnwidth}{!}
{\begin{tabular}{c|c c c|c c c}
\toprule
\multirow{2}{*}{ \bf Decoder Block } & \multicolumn{3}{|c|}{ \bf Extended MusicCaps} & \multicolumn{3}{c}{ \bf \ourdataset } \\
\cmidrule { 2 - 7 }
& \textbf{FAD} $\downarrow$ & \textbf{KL} $\downarrow$ & \textbf{FD} $\downarrow$ & \textbf{FAD} $\downarrow$ & \textbf{KL} $\downarrow$ & \textbf{FD} $\downarrow$ \\
\midrule
1 & 1.79 & 1.12 & 22.97 & 1.71 & 1.02 & 21.20 \\
1,2 & 1.53 & 1.05 & 22.76 & 1.27 & 0.86 & 20.93 \\
\CC{}\textbf{1,2,3} & \CC{}\textbf{1.12} & \CC{}\textbf{0.89} & \CC{}\textbf{22.65} & \CC{}\textbf{1.05} & \CC{}\textbf{0.72} & \CC{}\textbf{20.49} \\
\bottomrule
\end{tabular}}
\caption{Ablation of different decoder blocks}
\label{tab:decoder_block_ablation_rebuttal}
\end{table}

\subsection{On conditioning image} 

\modelname generates music from \underline{complementary information} from text and image modalities. While selecting images randomly, we have lower FAD/KL/FD scores of 6.38/1.73/26.45 and 8.33/1.57/28.64 on the extended MusicCaps and MeLBench datasets respectively, as it gets conditioned on random image semantics. We see similar trend in the baselines too, and \modelname still outperforms them. Retrieving or generating image from conditioning text, will also have similar effect due to semantic similarity in both conditioning domains.

\subsection{Alternate visual conditioning} 
We compare alternate conditioning from ViT features and ControlNet here. The semantics contained in these representations are inferior to those from text-to-image models (similar to findings in \cite{yang2023diffusion}). Further, our visual synapse effectively adapts them by learning to modulate the representations, specific to music synthesis. Moreover compared to the generalist model (that consumes multiple modalities) in AudioLDM2 \cite{audioldm2}, our specialist synaptic model generates better music. Also, their feature concatenation strategy is inferior to our visual synapse, as evident from Tab. \ref{tab:comparisons_rebuttal}.

\begin{table}[H]
\centering
\resizebox{\columnwidth}{!}
{\begin{tabular}{c|c c c|c c c}
\toprule
\multirow{2}{*}{ \bf Model } & \multicolumn{3}{|c|}{ \bf Extended MusicCaps} & \multicolumn{3}{c}{ \bf \ourdataset } \\
\cmidrule { 2 - 7 }
& \textbf{FAD} $\downarrow$ & \textbf{KL} $\downarrow$ & \textbf{FD} $\downarrow$ & \textbf{FAD} $\downarrow$ & \textbf{KL} $\downarrow$ & \textbf{FD} $\downarrow$ \\
\midrule
CLIP ViT Feats \cite{clip} & 1.83 & 1.15 & 23.03 & 1.77 & 1.04 & 21.48 \\
Control Net \cite{controlnet} & 1.65 & 1.09 & 22.94 & 1.25 & 0.85 & 20.91 \\
AudioLDM2 \cite{audioldm2} & 1.77 & 1.13 & 22.96 & 1.74 & 1.02 & 21.42 \\
\CC{}\textbf{Ours} & \CC{}\textbf{1.12} & \CC{}\textbf{0.89} & \CC{}\textbf{22.65} & \CC{}\textbf{1.05} & \CC{}\textbf{0.72} & \CC{}\textbf{20.49} \\
\bottomrule
\end{tabular}}
\caption{Comparison against different visual conditioning}
\label{tab:comparisons_rebuttal}
\end{table}

\subsection{Subjective analysis}

We complement our OVL scores with subjective descriptions, where we ask the evaluators to justify the score, stratify  them based on OVL scores, and report the most frequent reasons in Tab. \ref{tab:subjective_rebuttal}. 

\begin{table}[]
% \vspace{-20pt}
\small
\centering
\setlength{\tabcolsep}{5pt}
\renewcommand{\arraystretch}{0.25}
\resizebox{1.\columnwidth}{!}
{\begin{tabular}{c | l}
\toprule
\bf OVL Range & \bf  Reasons \\
\midrule
\multirow{2}{*}{0-25} & Discordant sound, unpleasant, poor quality, mismatched genre, not cohesive, \\
& repetitive melody, distractive background noise, unpleasant timbre, lack of contrast.  \\ \midrule
\multirow{2}{*}{26-50} & Unappealing instrumentation, lack of emotional resonance, unusual degree of dissonance, \\
& complex narrative, unrelatable theme, abrupt transition, unbalanced sound levels. \\ \midrule
\multirow{2}{*}{51-75} & Inconsistent mood, uninteresting chord progression, uneven transition between sections, \\
& has a nostalgic appeal, cinematic quality, spirituality. \\ \midrule
\multirow{3}{*}{76-100} & Exudes calmness,  cohesive, pleasing sequence of notes, well balanced combinations,  \\
& engaging rhythmic pattern, evoke a sense of groove, nice arrangement of instruments, \\
& strong sense of expression, authentic, vibrant texture, catchy, intuitive and natural flow.   \\
\bottomrule
\end{tabular}}
\caption{Subjective analysis on generated samples}
\label{tab:subjective_rebuttal}
\end{table}

\subsection{Learnable versus Fixed $\alpha$ Parameters}

\begin{table}[H]
\centering
\resizebox{\columnwidth}{!}
{\begin{tabular}{c|c c c|c c c}
\toprule 
\multirow{2}{*}{ \bf Fusion parameter $\alpha$} & \multicolumn{3}{|c|}{ \bf Extended MusicCaps} & \multicolumn{3}{c}{ \bf \ourdataset } \\
\cmidrule { 2 - 7 }
& \textbf{FAD} $\downarrow$ & \textbf{KL} $\downarrow$ & \textbf{IMSM} $\uparrow$ & \textbf{FAD} $\downarrow$ & \textbf{KL} $\downarrow$ & \textbf{IMSM} $\uparrow$ \\
% \midrule
% \multicolumn{8}{c}{\textit{Fixed $\alpha$ for all blocks.}}\\
\midrule
$\alpha=0$ & 3.07 & 1.21 & - & 3.11 & 1.19 & - \\
$\alpha=0.10$ & 2.98 & 1.17 & 0.51 & 3.03 & 1.07 & 0.56 \\
$\alpha=0.50$ & 1.17 & 0.93 & 0.71 & 1.12 & 0.79 & 0.77\\
$\alpha=0.90$ & 4.96 & 1.38 & 0.85 & 4.11 & 1.29 & 0.89 \\
$\alpha=1.0$ & 5.62 & 1.54 & - & 4.16 & 1.37 & - \\
\CC{}{\textbf{Learnable $\alpha$}} & \CC{}\textbf{1.12} & \CC{}\textbf{0.89} & \CC{}\textbf{0.76} & \CC{}\textbf{1.05} & \CC{}\textbf{0.72} & \CC{}\textbf{0.83} \\
\bottomrule
\end{tabular}}
\caption{
Analyzing the effect of having fixed versus learnable $\alpha$.
}
\label{tab:fixed_vs_learnable_alpha}
\end{table}
We study the impact when $\alpha$ is kept frozen as compared to being learnable here. The first five entries in \cref{tab:fixed_vs_learnable_alpha} denote the cases where the value of $\alpha$ is unaltered during training and kept constant at $0$, $0.10$, $0.50$, $0.90$, and $1.0$ respectively. Experimental results demonstrate that a learnable value of $\alpha$ produces significantly better results as compared to the fixed counterpart, as the model has the flexibility to learn them to effectively balance between both the conditioning modalities.

\begin{figure*}
    \centering
    \includegraphics[width=\textwidth]{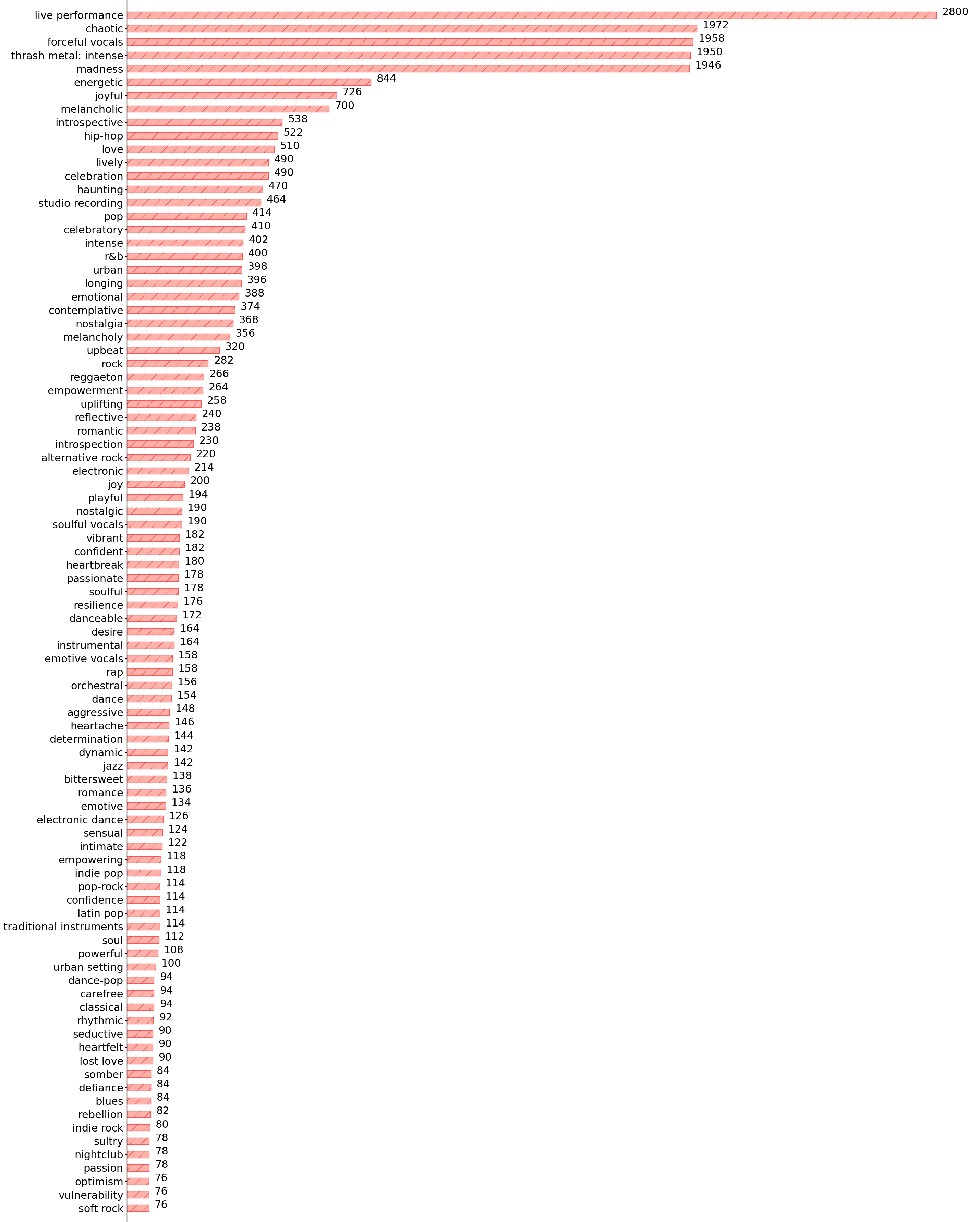}
    \caption{Frequency of top 90 words from \ourdataset}
    \label{fig:dataset_word_frequency}
\end{figure*}

\begin{table*}[!t]
\centering
\resizebox{\textwidth}{!}
{\begin{tabular}{l|c}
\toprule
% \rowcolor{Dark}
\textbf{Genre} & \textbf{Subgenre}\\
\midrule
\multirow{2}{*}{Hip-Hop} & \CCC{}{Alternative Hip Hop, Rap, Pop Rap, Trap, Melodic Rap, Gangster Rap, Southern Hip Hop, Urban Contemporary, Crunk, German Hip Hop, Rap Conscient, Italian Hip Hop,}\\
& \CCC{}{ East Coast Hip Hop, Hardcore Hip Hop, Atl Hip Hop, Dirty South Rap, Russian Hip Hop, Polish Trap, Underground Hip Hop, Funk Carioca, West Coast Rap, Cloud Rap}\\
\midrule
\multirow{2}{*}{Pop} & Dance Pop, Pov- Indie, Singer-Songwriter Pop, Mexican Pop, J-Pop, Latin Arena Pop, Indie Pop, Modern Country Pop, Art Pop, Alt Z, Indietronica,\\
&    New Wave Pop, Spanish Pop, Italian Adult Pop, Electropop, Turkish Pop, Reggae Fusion, Post-Teen Pop, Hip Pop, Ccm, Indonesian Pop, Pop Nacional \\
\midrule
\multirow{2}{*}{Latin} & \CCC{}{ Latin Pop, Trap Latino, Urbano Latino, Reggaeton, Musica Mexicana, Rock En Espanol, Norteno, Sierreno,R\&B Francais, Reggaeton Colombiano, Sad Sierreno, }\\
& \CCC{}{ Mpb, Sertanejo, Tropical, Latin Alternative, Banda, Corrido, Grupera, Ranchera, Trap Brasileiro, Rap Conciencia, Urbano Espanol  }\\
\midrule
\multirow{2}{*}{Electronic} & Edm, Pop Dance, Uk Dance, Electronica, Electro House, House, German Dance, Tropical House, Downtempo, Brostep, \\
&  Stutter House, Progressive House, Slap House, Big Room, Chill House, New French Touch, Dancefloor Dnb, Chillhop, Pop Edm, Lo-Fi Beats, Trance, Metropopolis  \\
\midrule
\multirow{2}{*}{R\&B} & \CCC{}{Soul, Indie Soul, Quiet Storm, Neo Soul, Funk, Alternative R\&B, Disco, Pop Soul, Afrobeats, Bedroom R\&B, Dark R\&B,}\\
& \CCC{}{ Reggae, British Soul, Contemporary R\&B, Hi-Nrg, Classic Soul, Uk Contemporary R\&B, Motown, New Jack Swing, Gospel, Roots Reggae, Philly Soul} \\
\midrule
\multirow{2}{*}{Easy listening} &  Adult Standards, Chanson, Soundtrack, Show Tunes, Hollywood, Movie Tunes, Cartoon, Japanese Soundtrack, Broadway, Deutsch Disney, Swing, British Soundtrack,  \\
&  Lounge, Preschool Children's Music, Scorecore, Romantico, Classic Girl Group, Children's Music, Electro Swing, French Soundtrack, French Movie Tunes, Classic Soundtrack    \\
\midrule
\multirow{2}{*}{World / traditional} & \CCC{}{Folkmusik, Modern Bollywood, Filmi, Pop Urbaine, World, Afroswing, Dancehall, World Worship, Entehno, Sufi, Naija Worship, Classic Bollywood,  }\\
& \CCC{}{ Nouvelle Chanson Francaise, Modern Reggae, Laiko, Classic Opm, Uk Dancehall, South African Pop Dance,  Chutney, Celtic,  Manila Sound, Azontobeats   }\\
\midrule
\multirow{2}{*}{Jazz} &  Vocal Jazz, Bossa Nova, Dinner Jazz, Contemporary Post-Bop, Jazz Fusion, Nu Jazz, Background Jazz, Smooth Jazz, Jazz Funk, Contemporary Vocal Jazz, Jazz Piano, \\
&   Jazztronica, Hard Bop, Smooth Saxophone, Cool Jazz, Nz Reggae, Soul Jazz, Torch Song, Folclore Salteno, Indie Jazz, Contemporary Jazz, Brazilian Jazz \\
\midrule
\multirow{2}{*}{Rock} & \CCC{}{ Permanent Wave, Modern Rock, Classic Rock, Mellow Gold, Album Rock, Soft Rock, Pop Rock, Alternative Rock, Hard Rock,}\\
& \CCC{}{Folk Rock, New Wave, New Romantic, Indie Rock, Heartland Rock, Latin Rock, Art Rock, Blues Rock, Dance Rock, Country Rock, Alternative Dance, Pop Punk, Punk  }\\
\midrule
\multirow{2}{*}{Classical} &  Orchestral Soundtrack, Compositional Ambient, Classical Performance, Javanese Dangdut, Italian orchestra, Orchestral Performance, Neo-Classical, Orchestra, Classical Piano, British Orchestra, Choral, \\
& Opera, Indian Classical, Hungarian Classical, Epicore, Impressionism,  Chamber Orchestra,  Historically Informed Performance, Violin, Baroque Ensemble, Symfonicky Orchestra, Japanese Guitar     \\
\midrule
\multirow{2}{*}{Blues} & \CCC{}{ Electric Blues, Jazz Blues, British Blues, Modern Blues, Malian Blues, Rebel Blues, Acoustic Blues, Rhythm And Blues, Doo-Wop, Traditional Blues, Soul Blues,  Louisiana Blues, }\\
& \CCC{}{ Garage Rock Revival, Indie Quebecois, New Orleans Blues, Texas Blues, Country Blues, Australian Garage Punk, Chicago Blues, Delta Blues, Memphis Blues, Slack-Key Guitar}\\
\midrule
\multirow{2}{*}{Metal} &  Alternative Metal, Post-Grunge, Nu Metal, Rap Metal,
 Groove Metal,
 Power Metal,
 Melodic Metalcore,
 Metalcore,
 Skate Punk
 
\\
& Glam Metal,
 Thrash Metal,
 Speed Metal,
 Death Metal,
 Funk Metal,
 Screamo,
 Nerdcore Brasileiro,
 Industrial Metal,
 Comic Metal,
 Symphonic Metal,
 Deathcore,
 Gothic Metal,
 Progressive Metal,\\
\midrule
\multirow{2}{*}{Country} & \CCC{}{Contemporary Country,
 Agronejo,
 Arrocha,
 Country Road,
 Sertanejo Universitario,
 Outlaw Country,
 Nashville Sound,
 Pop Rap Brasileiro,
 Pagode Novo,
 
}\\
& \CCC{}{Arrochadeira,
 Forro,
 Forro De Favela,
 Funk 150 Bpm,
 Progressive Bluegrass,
 Black Americana,
 Axe,
 Bandinhas,
 Funk Ostentacao,
 Alternative Country,
 Piseiro,
 Jam Band,
 Classic Texas Country}\\
\midrule
\multirow{2}{*}{Folk/ acoustic} & Singer-Songwriter,
 Neo Mellow,
 Indie Folk,
 New Americana,
 Stomp And Holler,
 British Singer-Songwriter,
 Melancholia,
 Lilith,
 Turbo Folk,
 Countrygaze,
 Neo-Psychedelic,
\\
& Pop Folk,
 Turkish Folk,
 Ambient Folk,
 Modern Indie Folk,
 Rune Folk,
 Indian Folk,
 Fantasy,
 Alternative Americana,
 Ska Punk,
 Vbs,
 German Indie
\\
\midrule
\multirow{2}{*}{New age} & \CCC{}{Rain,
 Color Noise,
 Sleep,
 Sound,
 Healing Hz,
 Solfeggio Product,
 Indie Game Soundtrack,
 Ocean,
 Environmental,
 Water,
 Piano Cover,
}\\
& \CCC{}{Acoustic Guitar Cover,
 Lullaby,
 High Vibe,
 Instrumental Worship,
 Atmosphere,
 Background Music,
 Ambient Worship,
 Binaural,
 Brain Waves,
 Background Piano,
Fourth World
}\\
\bottomrule
\end{tabular}}
\caption{Genre and sub-genre-wise division of the collected samples. Our dataset encompasses samples from 15 different genres each further divided into 22 sub-genres}
\label{tab:dataset_genre_division}
\end{table*}

\begin{figure*}
    \centering
    \includegraphics[width=\textwidth, height=\textheight]{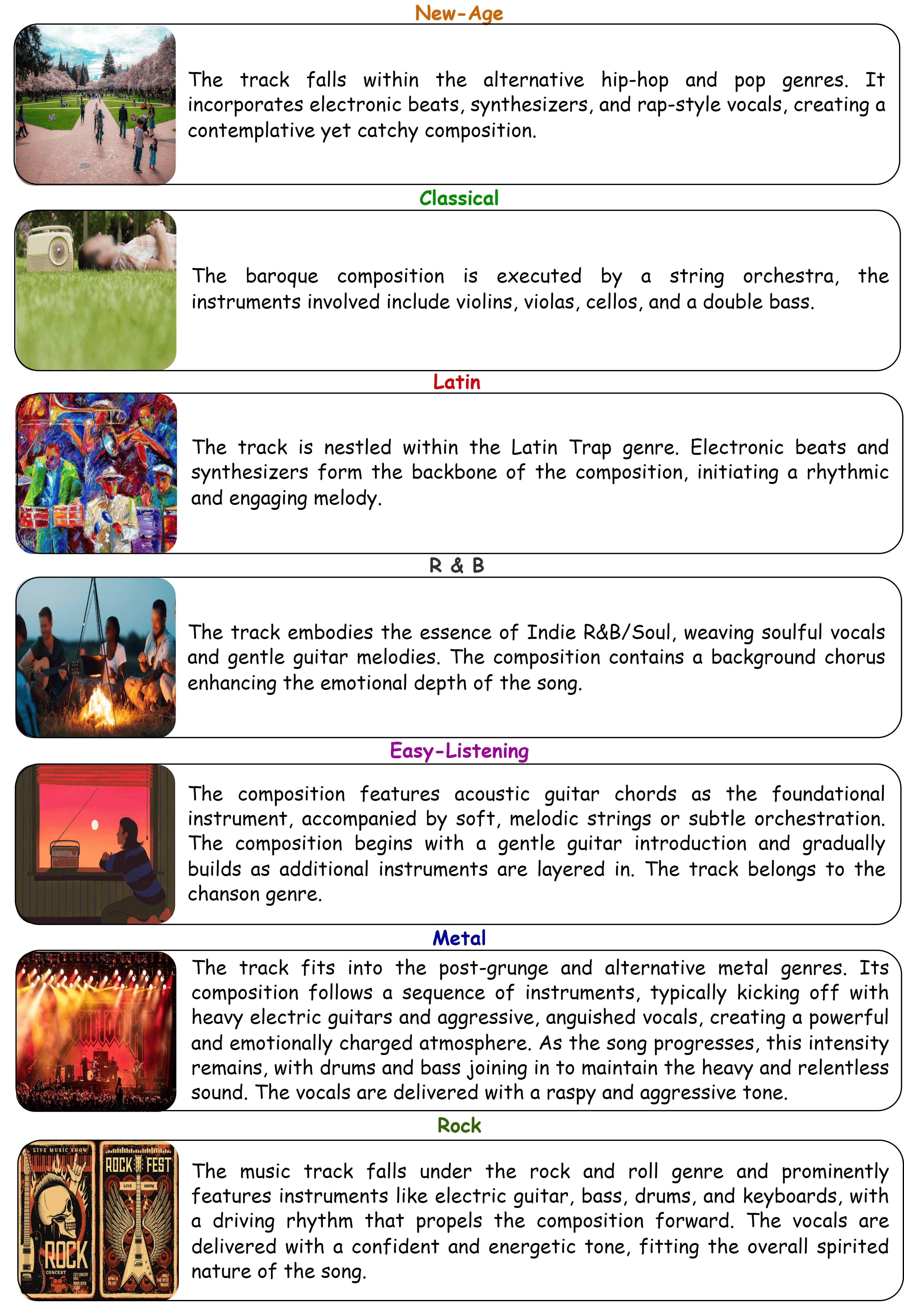}
    % \caption{Samples from \ourdataset}
    \label{fig:dataset2_1}
\end{figure*}

\begin{figure*}
    \centering
    \includegraphics[width=\textwidth]{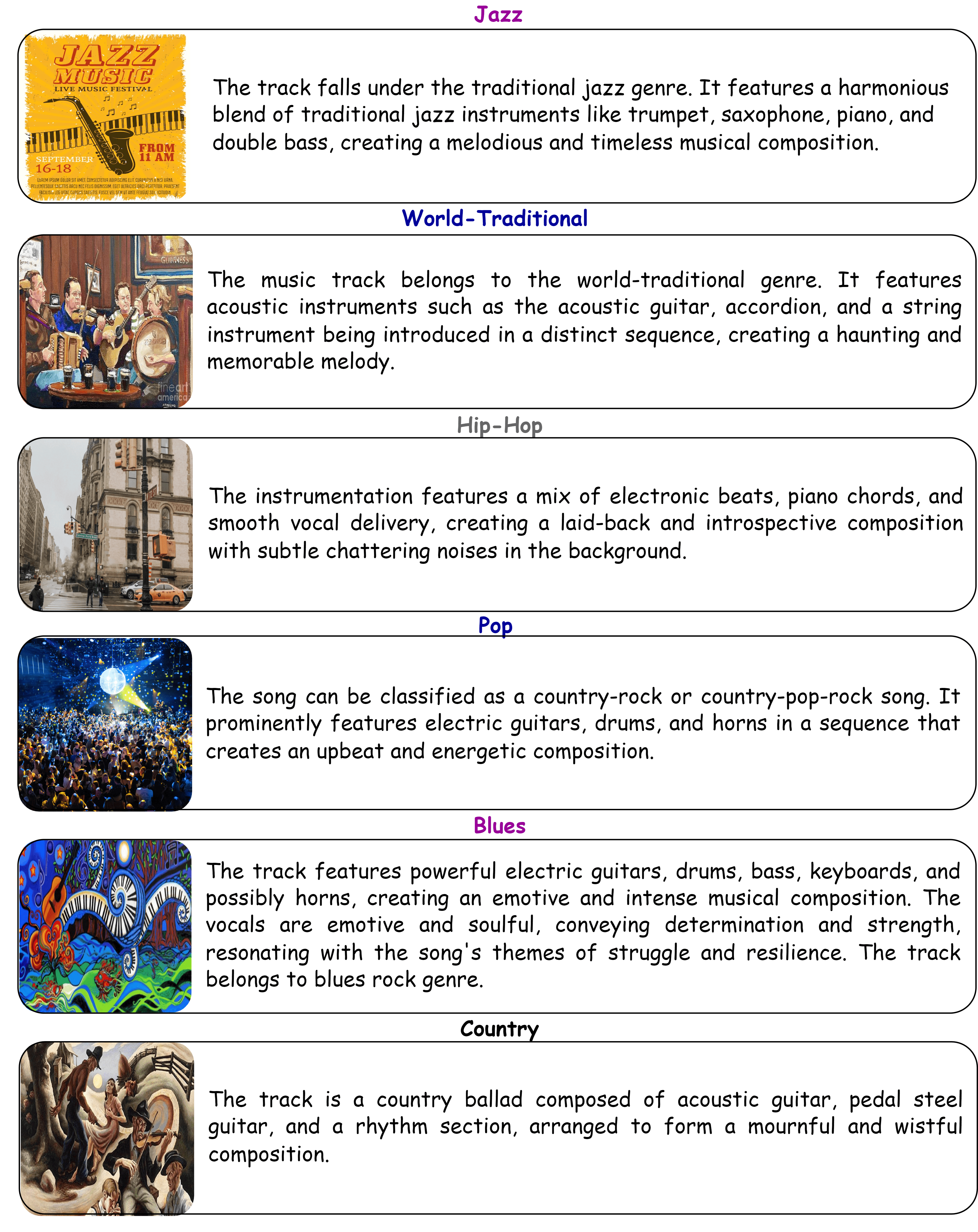}
    \caption{Samples from \ourdataset.}
    \label{fig:dataset_examples_supp}
\end{figure*}

\section{Dataset Details}
\label{sec: more dataset details}

\subsection {\ourdataset Statistics}
\begin{table}[H]
\centering
\resizebox{\columnwidth}{!}
{\begin{tabular}{c | c | c}
\toprule
\textbf{Type of image} & \textbf{\# Pieces} & \textbf{Percentage (\%) in Dataset}\\
\midrule
Natural image & 3206 & 28 \\
Animation & 2404 & 21 \\
Poster & 2748 & 24 \\
Painting / Sketch & 3092 & 27 \\
\bottomrule
\end{tabular}}
\caption{Image categories in \ourdataset.}
\label{tab:dataset}
\end{table}
\cref{tab:dataset} presents the distribution of the image samples in \ourdataset. To maintain a fair balance across different distributions we collect samples from 4 different categories: natural images, animations, posters, paintings/sketches. This ensures that \modelname is trained with ample examples from each of these classes and is equipped to tackle images from any of these very frequent and popular classes better. \ourdataset comprises \ourdatasetsize samples which is $\sim2$x larger than the next largest dataset MusicCaps \cite{musiclm}.

\cref{fig:dataset_word_frequency} presents the frequency of the top 90 words in \ourdataset. The annotators were asked to write free-form text descriptions of the musical pieces with an emphasis on the musicality of the samples. We observe that the annotation contains important cues about the nature of the audio track (e.g., `live performance', `chaotic', `forceful vocals', etc). These can supplement a model with useful pieces of information regarding the aesthetics of the composition.

\subsection {Dataset Hierarchy and Samples}
\cref{tab:dataset_genre_division} contains the genre and sub-genre-wise division of the samples collected in \ourdataset. We categorise the collected musical samples into 15 broad categories with each of them having 22 sub-genres to facilitate fine-grained control over the composition through the image (theme) and text-instructions (details on musicality). The samples are divided across different genres roughly equally to maintain a good balance.

\cref{fig:dataset_examples_supp} presents one sample from each of the remaining 13 categories (Electronic and Folk Acoustic present in the main paper). As can be seen from the examples, the captions are of varied lengths and the images are from different distributions (natural images, animation, paintings, etc.).

\subsection{Extended MusicCaps Data Collection}
MusicCaps \cite{musiclm} is a music caption dataset comprising music clips from AudioSet \cite{gemmeke2017audio} paired with corresponding text descriptions in English. The collection consists of a total of 5,521 examples, out of which 2,858 are from the AudioSet eval and 2,663 are from the AudioSet train split. The authors further tag 1,000 samples as a balanced subset of the dataset - equally divided across genres. All examples in the balanced subset are from the AudioSet eval split. As our setup is not restricted to text and requires joint conditioning in the form of images as well, we supplement this dataset by collecting 2 carefully chosen image frames for each of the 10-second samples from the corresponding YouTube video or web. As some of the samples are not live anymore, we were able to collect a total of \musiccapssamples~ samples which we divided into a 60\%/20\%/20\% split for train/validation/test respectively.

\begin{figure}
    \centering
    \includegraphics[width=0.47\textwidth]{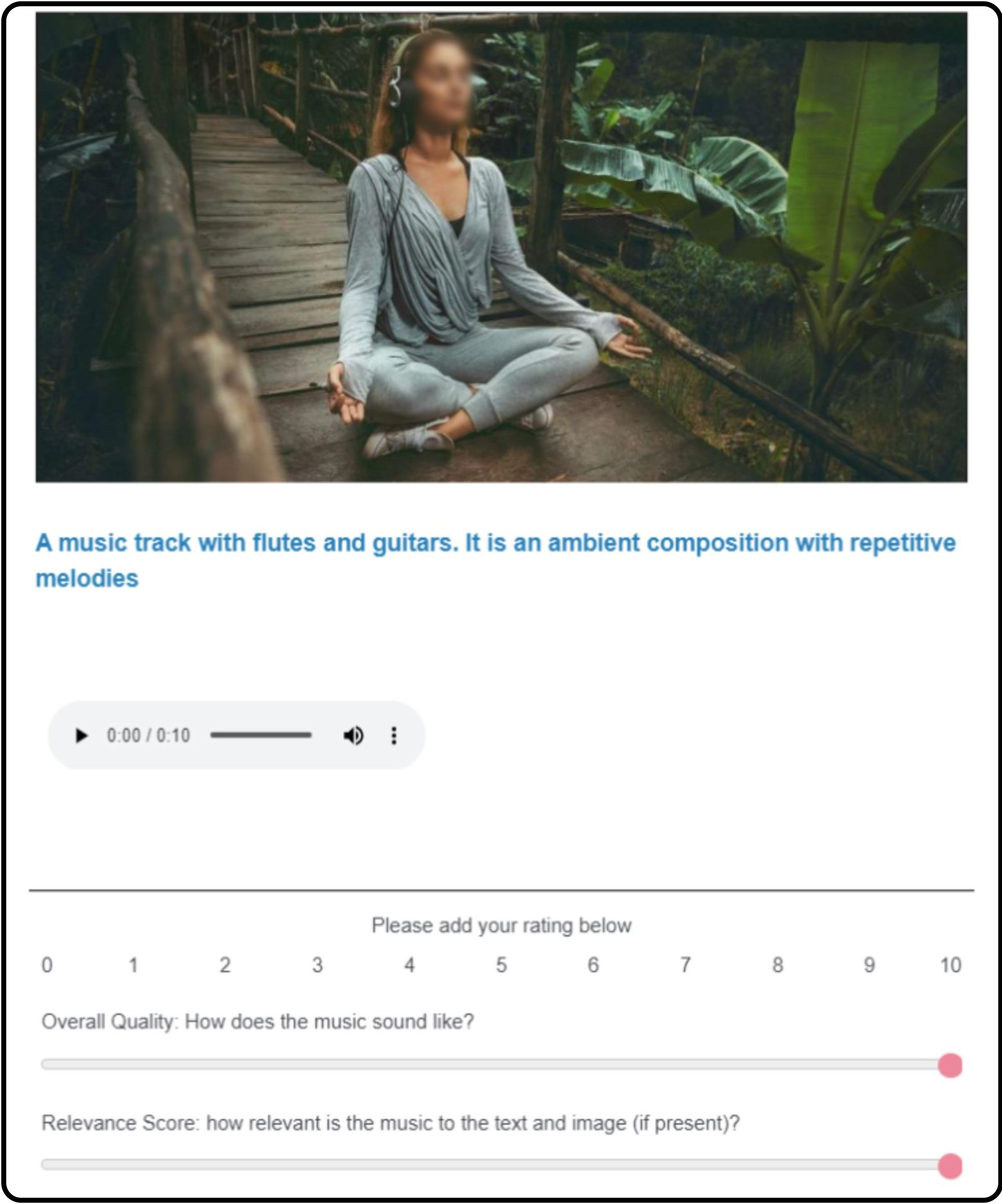}
    \caption{User study interface to collect OVL and REL scores. }
    \label{fig:user-study_ovl_rel}
\end{figure}

\begin{figure}[!t]
    \centering
    \includegraphics[width=0.47\textwidth]{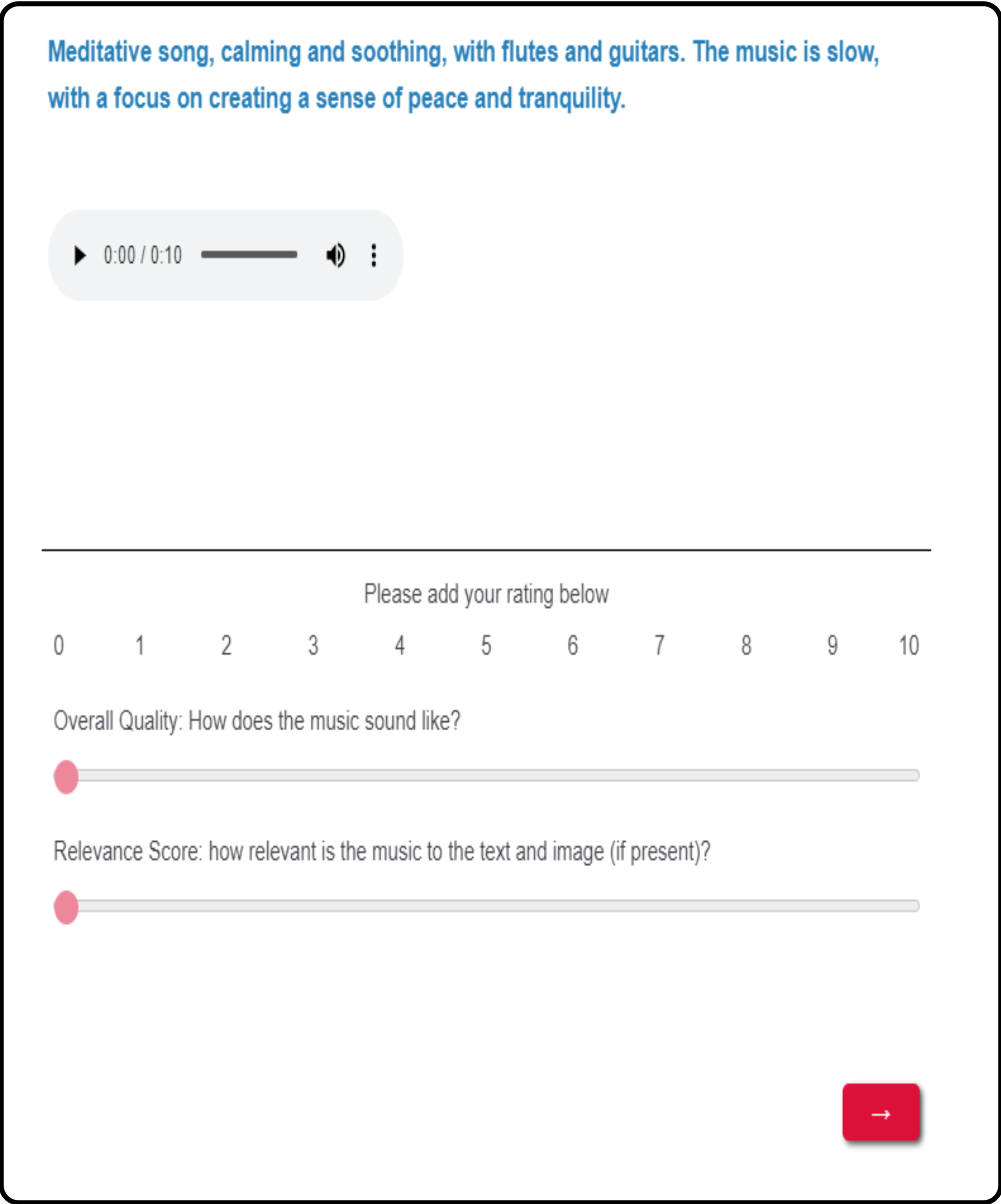}
    \caption{User study interface for comparison against prior text-to-music methods}
    \label{fig:user-study-comparison}
\end{figure}

\begin{figure}
    \centering
    \includegraphics[width=0.47\textwidth]{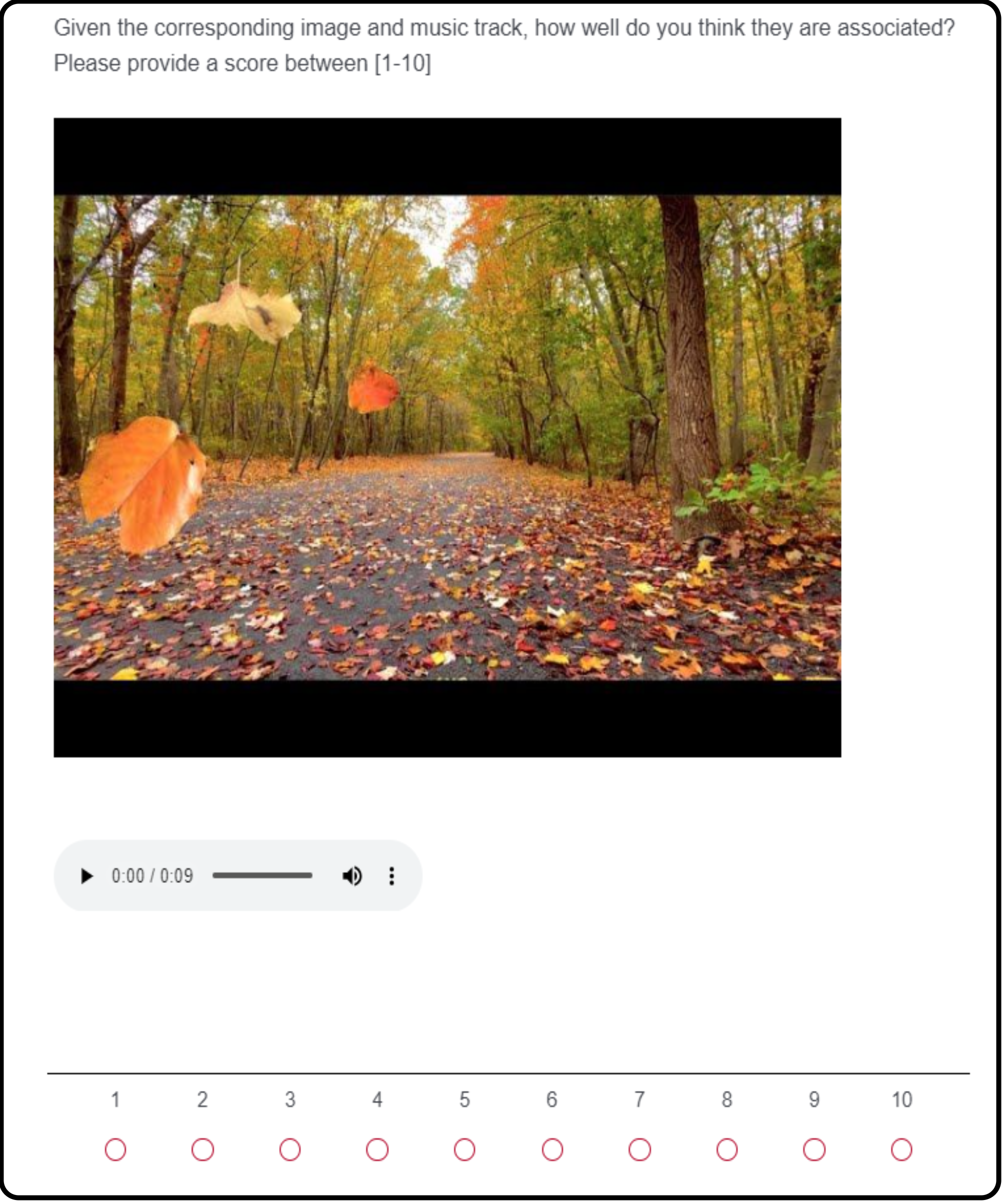}
    \caption{User study interface to obtain IMSM scores}
    \label{fig:user-study_imsm}
\end{figure}

\section{User Study Details}
\label{sec: more user study}

% \subsection{Details on user study}
\cref{fig:user-study_ovl_rel} presents the user study interface. To obtain the OVL and REL scores, we provide the participants with an image-text pair and the audio sample generated by \modelname. For the overall audio quality score (OVL) the participants are instructed to add their score between [1,10] while for the relevance score (REL), they are required to rate the sample based on its similarity with the input image-text pairs. 

In \cref{fig:user-study-comparison} we compare our method against prior text-to-music methods and report the OVL and REL scores in the main paper (Tab. 1). In this case, the participants were presented with only the text-music pairs.

\cref{fig:user-study_imsm} shows the user study interface for the IMSM score. For this, the participants were presented with image-music pairs and asked to provide their rating between [1,10], with 1 being the lowest. The higher the score, the more perceptually similar the participant has found the image-music pair to be.

\section{Inspiration from Conditional Image Generation}
\label{sec: conditional image gen}

Powered by architectural improvements and the availability of large-scale, high-quality paired training data, conditional image generation methods have made considerable progress in the generative AI space. Promising results from transformer-based auto-regressive approaches \cite{yu2022scaling, ramesh2021zero} were boosted by diffusion model-based methods \cite{rombach2022high, saharia2022photorealistic, nichol2021glide}. These approaches have been naturally extended to generate videos from text prompts too \cite{singer2022make, wu2023tune, ho2022imagen}. Latent diffusion models \cite{rombach2022high} do the diffusion process in the latent space of a pre-trained VQ-VAE \cite{van2017neural}. This significantly reduced the compute requirements when compared with image diffusion methods. \citet{classifierfree} proposed classifier-free guidance to enhance image quality.
Text-to-music and text-to-audio methods are heavily inspired by the success of text-to-image generative methods, and so are we.

\section{Related Audio Concepts} 
\label{sec: related audio concepts}
% VAE
% \noindent{\textbf{Audio VAEs:}}
The Multimodal Variational Auto-encoders (MVAEs) are latent variable generative models to learn more generalizable representations from diverse modalities through joint distribution estimation. \citet{arik2018neural} pioneered a neural audio synthesis model based on VAEs. Their approach demonstrated promising results in generating realistic audio samples by learning a latent representation of the audio data. Inspired by this VAEs have been widely used in the audio processing domain for speech synthesis \cite{liu2021vara, zhang2019learning, tan2024naturalspeech}, audio generation \cite{jiang2020transformer, tango, caillon2021rave}, and audio denoising \cite{sadeghi2020audio, bando2020adaptive}. 

% vocoder
% \noindent{\textbf{Vocoders:}}
Vocoders are used for a variety of purposes across different domains due to their ability to manipulate and synthesize audio signals efficiently. Among other prominent applications of vocoder, neural voice cloning \cite{arik2018neural, jia2018transfer}, voice conversion \cite{liu2018wavenet}, and speech-to-speech synthesis \cite{jia1904direct} are very popular. GAN-based vocoders \cite{kong2020hifi} have been employed to generate high-fidelity raw audio conditioned on mel spectrogram. More recently, WaveRNN \cite{kalchbrenner2018efficient} has been applied for universal vocoding task \cite{lorenzo2018towards, paul2020speaker, jiao2021universal}.

%spectrograms
% \noindent{\textbf{Spectrograms:}}
Spectrograms are a powerful tool for analyzing time-varying signals such as audio and speech. They provide a visual representation of the frequency content of a signal over time, making them widely used in speech processing \cite{michelsanti2021overview, chuang2022improved, seo2023avformer}, music analysis \cite{melody, mousai}, and audio synthesis \cite{adverb, makeanaudio, tango, audit, audioldm} in general. Audio spectrograms are also massively deployed in different audio visual applications \cite{adverb, park2024can, sung2023sound}.

\vspace{0.2in}
\noindent{\textbf{Acknowledgements:}} We would like to sincerely thank the data annotators and the volunteers who took part in the user study. We would also like to extend our gratitude to the anonymous reviewers for their constructive and thoughtful feedbacks.

% {
%     \small
%     \bibliographystyle{ieeenat_fullname}
%     \bibliography{main}
% } 

\end{document}